\documentclass[journal]{IEEEtran}
\usepackage{amsmath,amsfonts}
\usepackage{algorithmic}
\usepackage{algorithm}
\usepackage{array}
\usepackage{caption} 
\usepackage{tabularx}
\usepackage{multirow}
\usepackage{tikz}
\usetikzlibrary{shapes.geometric, arrows, positioning}
\usepackage[caption=false,font=normalsize,labelfont=sf,textfont=sf]{subfig}
\usepackage{textcomp}
\usepackage{stfloats}
\usepackage{url}
\usepackage{verbatim}
\usepackage{graphicx}
\usepackage{epstopdf}
\usepackage[numbers,sort&compress]{natbib}
\def\BibTeX{{\rm B\kern-.05em{\sc i\kern-.025em b}\kern-.08em
    T\kern-.1667em\lower.7ex\hbox{E}\kern-.125emX}}
\usepackage{balance}
\begin{document}
\title{Strategies for decentralised UAV-based collisions monitoring in rugby }
\author{Yu Cheng and Harun Šiljak,~\IEEEmembership{Senior Member,~IEEE}

\thanks{The authors are with the School of Engineering, Trinity College Dublin, Ireland.
This paper is partly supported by Research
Ireland, the European Regional Development Fund (Grant No. 13/RC/2077\_P2) and the EU MSCA Project ”COALESCE” (Grant No. 101130739)

Corresponding author harun.siljak@tcd.ie}}

\markboth{Journal of \LaTeX\ Class Files,~Vol.~18, No.~9, September~2020}%
{How to Use the IEEEtran \LaTeX \ Templates}

\maketitle

\begin{abstract}
Recent advancements in unmanned aerial vehicle (UAV) technology have opened new avenues for dynamic data collection in challenging environments, such as sports fields during fast-paced sports action. For the purposes of monitoring sport events for dangerous injuries, we envision a coordinated UAV fleet designed to capture high-quality, multi-view video footage of collision events in real-time. The extracted video data is crucial for analyzing athletes' motions and investigating the probability of sports-related traumatic brain injuries (TBI) during impacts. This research implemented a UAV fleet system on the NetLogo platform, utilizing custom collision detection algorithms to compare against traditional TV-coverage strategies. Our system supports decentralized data capture and autonomous processing, providing resilience in the rapidly evolving dynamics of sports collisions.

The collaboration algorithm integrates both shared and local data to generate multi-step analyses aimed at determining the efficacy of custom methods in enhancing the accuracy of TBI prediction models. Missions are simulated in real-time within a two-dimensional model, focusing on the strategic capture of collision events that could lead to TBI, while considering operational constraints such as rapid UAV maneuvering and optimal positioning. Preliminary results from the NetLogo simulations suggest that custom collision detection methods offer superior performance over standard TV-coverage strategies by enabling more precise and timely data capture. This comparative analysis highlights the advantages of tailored algorithmic approaches in critical sports safety applications.
\end{abstract}


\section{Background Introduction}
Collisions are inherent in contact sports, leading to an elevated risk of traumatic brain injuries (TBIs), particularly in high-impact sports such as rugby, where head collisions are a primary concern. Recent studies indicate that a significant proportion of rugby-related injuries involve the head. For instance, a systematic review by Paul et al. \cite{tackles} found that rugby union matches typically feature an average of 156 tackles per match, compared to 14 tackles per match in rugby sevens. Forwards generally experience more severe and heavier impacts than backs due to their involvement in high-impact collisions. Tucker et al. \cite{tackle01} demonstrated that $76\%$ of head injury assessments (HIA) in professional rugby occur during tackles, with tacklers facing a significantly higher risk of head injury compared to ball carriers. Similarly, Bathgate et al. \cite{tackle02} highlighted that head injuries, including concussions and lacerations, accounted for $25.1\%$ of all injuries among elite Australian rugby union players, with most injuries occurring during tackles. The combination of frequent tackles and player position significantly contributes to the elevated risk of TBIs, particularly for forwards, raising concerns not only about the immediate impact but also the long-term effects of head injuries. Repetitive TBIs, including concussions, can lead to severe neurodegenerative conditions such as chronic traumatic encephalopathy (CTE) and other related disorders \cite{longterm1,longterm2,longterm3,295}. Previous research by Rafferty et al. \cite{Rafferty} further emphasized that players become particularly vulnerable to head injuries after participating in 25 matches, with each subsequent concussion increasing the risk of future injuries by $38\%$. This growing body of evidence highlights the need for heightened awareness and preventive measures to mitigate the long-term risks associated with repeated head trauma in rugby players.

Frequent head injuries in rugby have prompted significant efforts to reduce their occurrence and severity, focusing on both immediate and long-term impacts of traumatic brain injuries (TBIs). Various strategies, such as the development of standardized assessment tools, wearable technologies, and advanced filtering methods, have been explored. Tools like the Sport Concussion Assessment Tool 6 (SCAT6) \cite{SCAT6} and World Rugby Head Injury Assessment (HIA01) aid \cite{Fuller} in evaluating concussions, while wearable devices and advanced impact analysis techniques show potential for enhancing the accuracy of concussion detection and prevention\cite{Celik,O'Connor,Sandmo,Lin,Gabler}. Additionally, recent research has leveraged deep learning for detecting high-risk tackles from match videos, showcasing advancements in preventive measures against TBIs in rugby \cite{Nonaka}.


Despite the advancements in concussion assessment, filtering techniques, and injury detection, several gaps remain in current solutions. Traditional tools, while effective, rely heavily on manual input, which introduces subjectivity and potential inaccuracies. Wearable technologies and filtering techniques, though promising, still face challenges with sensitivity, real-time accuracy, and false positives in natural game conditions. Moreover, many studies lack observer or video confirmation to validate recorded impacts, as highlighted by Patton et al. \cite{Patton}, leading to potential overestimation of head impact exposure. Deep learning systems like Nonaka et al.'s high-risk tackle detection offer new insights but struggle with practical implementation issues, such as processing speed and occlusion handling. Furthermore, most current systems focus on post-impact analysis, lacking proactive monitoring methods that can predict or prevent dangerous situations before they occur. While inertial measurement units (IMUs) offer a means to capture 3D kinematics over a large area without fixed installations, they often fall short in delivering the precision required for sport-specific analysis. Although modern IMUs are compact and capable of providing general kinematic data, attaching them to athletes can interfere with natural performance and equipment. Moreover, IMUs are susceptible to impact-related issues—potentially detaching during high-intensity collisions—and the raw data they generate is challenging to interpret accurately in relation to an athlete’s true movements \cite{imu2019}. This lack of precision and the intrusive nature of IMUs underscore the need for alternative, non-contact methods for detailed kinematic monitoring in dynamic sports environments. These gaps indicate a clear need for innovative solutions that can provide real-time, accurate, and decentralized monitoring in dynamic environments like rugby.


In response to these gaps, our contribution presents a smart, decentralized approach to monitoring collisions in rugby using UAVs. We designed a novel system that integrates UAV-based monitoring strategies with two-dimensional simulations using the NETLOGO platform. This system is capable of real-time tracking of player movements and potential collision risks, enhancing the overall safety and decision-making process during matches. Our decentralized design allows multiple UAVs to operate autonomously, monitoring the field from different angles and sharing data in a coordinated manner without relying on a central control unit.

The key innovation in our approach lies in the decentralized collision monitoring strategies, which enable the UAVs to collaborate and adjust their positioning dynamically based on player movement patterns. By analyzing real-time player data, our system can identify potential high-risk tackles before they occur, providing early warnings to sideline officials and medical teams. Furthermore, our UAV-based system reduces the limitations posed by occlusions and slow processing speeds in video-based systems, as UAVs can reposition themselves to maintain an optimal view of player interactions.

The rest of the paper is organized as follows: In Section II, we review related works, including existing camera-based systems, UAV monitoring strategies, and collision detection approaches in sports scenarios. Section III describes our detailed simulation framework, consisting of the Rugby Model and Drone Model, which defines player behaviors, game dynamics, and various UAV operational strategies. In Section IV, we present comprehensive simulation experiments and evaluate the performance of the proposed UAV-based collision detection strategies under varying fleet sizes, flight speeds, and detection radius. Finally, conclusions are drawn and future research directions are discussed in Section V.

\section{Related work}
Over the past decade, the field of sports analytics and live event broadcasting has evolved from static, fixed-camera systems to dynamic, UAV-based platforms that capture sports scenes from novel perspectives. Early work\cite{2007thomas,ALeman,REN} focused on using field markings and multi-camera setups to estimate camera poses and generate 3D reconstructions for tactical analysis. For instance, Alemán-Flores et al. \cite{ALeman} and Ren et al. \cite{REN}'s initial studies demonstrated that precise overlays and 3D game reconstructions could be achieved by exploiting natural field features and multi-view data fusion.

More recent research has increasingly leveraged the mobility of drones to overcome the limitations of fixed cameras. Z. Hong \cite{freeviewhong} introduced a monocular drone-based system that orbits an athlete to capture a full 360° view of an outdoor sports scene. By integrating structure-from-motion with neural rendering, their approach reconstructs both the dynamic athlete and its environment, providing a cost-effective alternative to conventional multi-camera arrays. This free-viewpoint video method enables dynamic scene replay from any angle, a significant advancement for real-time event analysis.

Simultaneously, advancements in motion capture have played a critical role in performance analysis and injury prevention. Ho et al. \cite{ho3D} developed a multi-UAV system that estimates 3D human pose in outdoor settings by coordinating multiple drones to maximize viewpoint diversity and minimize occlusions. In parallel, Jacobsson et al. \cite{jacapture} demonstrated a UAV-mounted depth camera system for markerless motion capture, showing that real-time skeleton tracking is feasible in field environments despite challenges such as limited flight endurance and sensor range constraints. Together, these studies indicate that UAV-based motion capture can provide high-fidelity data for analyzing player biomechanics and assessing injury risks.

In the realm of autonomous filming, Alcántara et al. \cite{Alcantara} proposed a system in which multiple drones collaboratively execute complex aerial shots. Their framework incorporates a high-level planning interface and distributed onboard controllers, enabling real-time, synchronized filming during live sports events. This autonomous multi-drone cinematography approach not only enhances coverage but also reduces the need for extensive human operation.

The design of the UAV platforms themselves is another important aspect. Casazola et al. \cite{casazola} presented a comprehensive study on UAV design for aerial filming, addressing issues such as stabilization, payload constraints, and flight endurance. Their work provides practical insights into building low-cost yet effective drones tailored for sports broadcasting, highlighting the trade-offs between agility and video quality.

In addition to these advances, sports such as rugby present unique challenges. In rugby events, multiple players often overlap, and even if camera occlusion issues are partially resolved, the competition among groups makes it difficult to fully meet the requirements of dynamic event coverage. UAV teams, however, offer a promising solution to these challenges. Moreover, for high-speed and high-participation sports like rugby, there is currently no comprehensive, advanced technology addressing head collision detection. To the best of our knowledge, our work is the first to propose such an approach. Existing two-dimensional agent-based modeling efforts for soccer \cite{noauthor_soccer_nodate, vainigli_lorenzovnglagent-based-football_2024} have laid a foundation. Our research significantly extends these efforts by not only simulating rugby and athlete dynamics but also by incorporating diverse strategy modifications and simulations for UAV swarm behavior.

Finally, the aesthetic and communicative potential of drones has also been explored. Hebbel-Seeger et al. \cite{hebbel} investigated how drone footage can enrich live sports broadcasting by offering immersive, bird’s-eye views. Their findings demonstrate that while aerial perspectives significantly enhance viewer engagement, challenges such as regulatory constraints and privacy concerns must be carefully managed.

In summary, these years have seen a clear evolution from static camera systems to agile UAV-based platforms capable of capturing free-viewpoint video, performing markerless motion capture, and autonomously filming dynamic sports events. Despite these advances, challenges related to real-time processing, flight endurance, and safety persist. Our work builds on these advancements by proposing a decentralized UAV fleet for real-time collision monitoring and injury assessment in high-impact sports, aiming to enhance both data accuracy and operational resilience.

\section{System Overview}

\begin{figure}[h]
\centering
\includegraphics[width=3.5in]{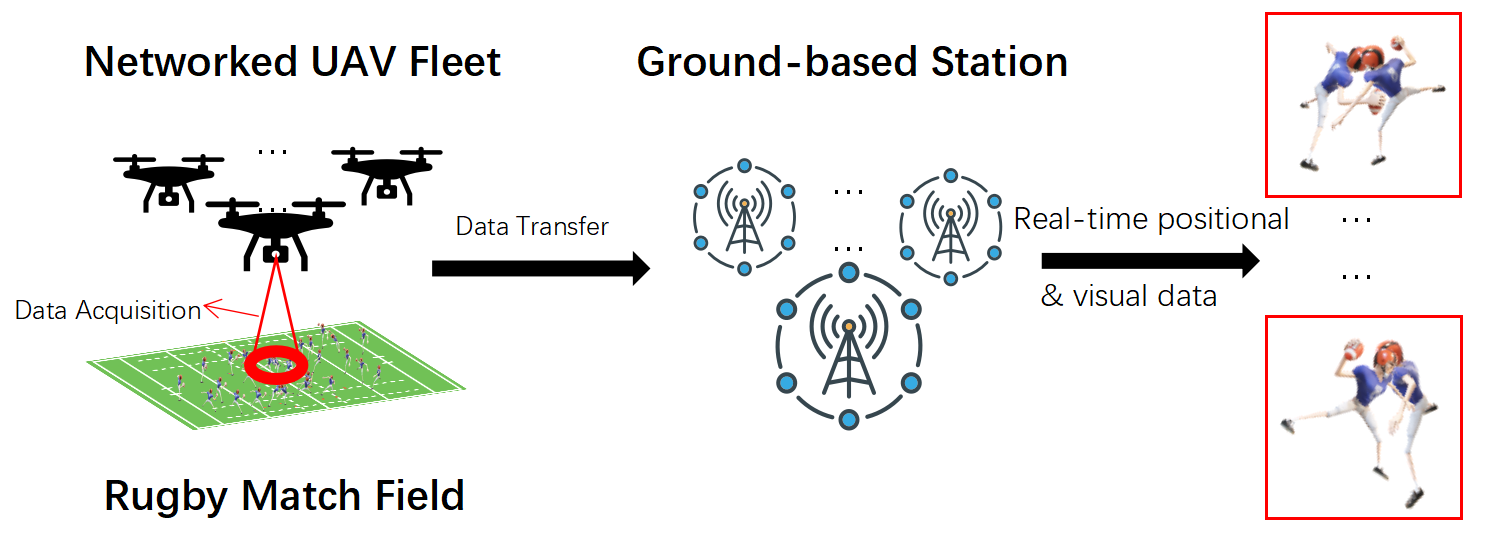}
\caption{The general framework of our model}
\label{O1}
\end{figure}

Building upon existing research, this paper introduces a decentralized UAV-based monitoring framework explicitly designed for head collision detection and analysis in rugby games. The proposed system's overall architecture is depicted in Fig. \ref{O1}.

The core components and their roles within the proposed framework are detailed as follows:  

\begin{itemize} 
\item \textbf{Networked UAV Fleet}: Multiple UAVs are strategically deployed above the rugby field, each equipped with high-precision GPS sensors and high-definition cameras. This configuration ensures comprehensive coverage, enabling detailed data acquisition for real-time tracking of player movements and potential collision events.

\item \textbf{Data Acquisition and Collision Monitoring Strategies}: UAVs actively monitor the rugby match environment by capturing high-quality visual data. The UAV fleet employs decentralized strategies, where each UAV independently identifies and tracks player interactions and potential collisions. These decentralized strategies are developed and rigorously evaluated using agent-based modeling and simulation via the NetLogo platform. Simulation results directly inform the performance and effectiveness of the decentralized collision monitoring algorithms.

\item \textbf{Simulation-based Data Analysis}: The proposed framework leverages agent-based simulations in NetLogo to replicate rugby game scenarios and UAV swarm behaviors. Data derived from these simulations is systematically analyzed to validate and refine collision detection strategies. This simulation-driven approach effectively complements physical testing by providing extensive scenario coverage and enhanced analytical precision.

\item \textbf{Advanced Data Processing and Analysis}: Following the decentralized data acquisition phase, advanced image processing, computer vision algorithms, and machine learning techniques are utilized to accurately identify collision events, assess injury risk, and extract performance-related insights from both real-world and simulated datasets. 

\end{itemize}

\section{Model Description}

This section outlines the simulation framework used to model a rugby game integrated with drone interactions. The simulation is implemented using the NETLOGO platform, which allows for agent-based modeling of complex systems. The framework is divided into two primary models: {\bf {the Rugby Model} } and {\bf {the Drone Model}}. The Rugby Model simulates the players, ball dynamics, and game environment, while the Drone Model introduces drones with various behaviors to interact within the game setting.

\begin{table}[h]
\centering
\caption{Table of Notation}
\setlength{\tabcolsep}{3.5pt}
\renewcommand{\arraystretch}{1.2}
\begin{tabular}{|c|p{3cm}|c|p{3cm}|}
\hline
\textbf{Symbol} & \textbf{Description} & \textbf{Symbol} & \textbf{Description} \\ \hline
$N$ or $n$ & Total number of drones & $R$ & Formation radius around target (ball/player) \\ \hline
$P_{\text{in}}$ & Set of players in competition with the ball & $P_{\text{out}}$ & Set of nearby players not in competition \\ \hline
$d_{\text{in}}$ & Distance threshold for $P_{\text{in}}$ & $d_{\text{out}}$ & Distance threshold for $P_{\text{out}}$ \\ \hline
$d_{\text{min}}$ & Minimum cumulative distance to the ball and teammate & $p_{\text{hr}}$ & High-risk player \\ \hline
$p_i$ & A player in $P_{\text{in}}$ & $p_j$ & Nearest teammate in $P_{\text{out}}$ \\ \hline
$\theta_{\text{step}}$ & Angle step for drone placement in circular formation & $\theta_i$ & Angle for positioning drone $d_i$ \\ \hline
$(x_i, y_i)$ & Position of drone $d_i$ & $(x_b, y_b)$ & Position of the ball \\ \hline
$x_{p_{\text{hr}}}, y_{p_{\text{hr}}}$ & Coordinates of the high-risk player $p_{\text{hr}}$ & $x_{C_j}, y_{C_j}$ & Centroid coordinates of cluster $C_j$ \\ \hline
$\rho_j$ & Density of cluster $C_j$ & $D_j$ & Number of drones assigned to cluster $C_j$ \\ \hline
$v_{\text{max}}$ & Maximum speed of drones & $d_{\text{safe}}$ & Minimum safe distance to avoid collisions \\ \hline
$(x_{\text{target}_i}, y_{\text{target}_i})$ & Target location for drone $d_i$ & $\mathbf{D}_i$ & Direction vector of drone $d_i$ \\ \hline
$\mathbf{R}_i$ & Repulsion vector to avoid collisions & $\mathbf{V}_i$ & Updated velocity of drone $d_i$ \\ \hline
$(x_i, y_i)$ & Current position of drone $d_i$ & $(x_j, y_j)$ & Position of neighboring drone $d_j$ \\ \hline
$C$ & Set of player clusters & $K$ & Total number of clusters \\ \hline
$r$ & Radius of drone coverage or density detection & $\Delta t$ & Simulation time step \\ \hline
$d_{ij}$ & Distance between drones $d_i$ and $d_j$ & $\mathbf{r}_{ij}$ & Repulsion force between $d_i$ and $d_j$ \\ \hline
\end{tabular}

\label{table_symbols}
\end{table}

\subsection{ \bf {The Rugby Model}}
The Rugby Model is designed to replicate the dynamics of a rugby game, including the field setup, player attributes, and ball mechanics. The simulation environment is initialized to represent a standard rugby field, and players are assigned roles and attributes to mimic real-world rugby scenarios.

\subsubsection{ \bf {Field Setup}}
The rugby field is configured to standard dimensions, with a width of 100 meters and a height of 70 meters, corresponding to the NETLOGO coordinates of 'width = 50' and 'height = 35'. The field is visually represented with green patches, and white lines are drawn to indicate boundaries and key field markings, such as the halfway line, try lines, and goal lines. The field setup includes:

- {\bf{Goal Areas}}: Defined at both ends of the field with patches representing the red and blue goals.
\

- {\bf{Field Lines}}: Thick solid white lines mark the center, sides, and try lines, while thin dashed lines indicate the 10-meter and 22-meter lines.

\subsubsection{ \bf {Players Initialization}}
This model simulates the movements and interactions of players on a rugby field, capturing the intricate dynamics of sports collisions. Each player is modeled as an autonomous agent with specific behaviors designed to mimic real-world actions, such as running, passing, and evading. The simulation adheres to the rules of rugby, including aspects like the kickoff by one team and the rule that passes must be backwards.

Players are created and assigned to two teams: red and blue. Each team consists of 15 players, reflecting standard rugby union team sizes. Players are further categorized based on their roles and attributes:

\begin{table}[h]
\centering
\caption{Player Roles and Attributes}
\begin{tabular}{|l|l|p{4cm}|}
\hline
\textbf{Category} & \textbf{Variable Name} & \textbf{Description and Values} \\
\hline
\multicolumn{3}{|c|}{\textbf{Roles}} \\
\hline
Role Type & --- & Players are assigned as \textbf{Defenders} or \textbf{Attackers}, and designated as \textbf{Team Players} or \textbf{Selfish Players}. \\
\hline
\multicolumn{3}{|c|}{\textbf{Attributes}} \\
\hline
Team Player & \texttt{teamplayer?} & Indicates if a player prefers to pass the ball to teammates. \\
& & \textbf{Possible Values}: \texttt{True} (Team Player), \texttt{False} (Selfish Player) \\
\hline
Defensive & \texttt{defensive?} & Determines if a player primarily focuses on defensive actions. \\
& & \textbf{Possible Values}: \texttt{True} (Defensive Player), \texttt{False} (Attacking Player) \\
\hline
Holding Ball & \texttt{holding-ball?} & Tracks if a player is in possession of the ball. \\
& & \textbf{Possible Values}: \texttt{True} (Has Ball), \texttt{False} (Does Not Have Ball) \\
\hline
Initial Position & --- & Players are positioned on the field based on predefined formations specific to their team and role. \\
& & \textbf{Values}: Coordinates (\texttt{x}, \texttt{y}) on the field \\
\hline
Run Speed & \texttt{run-speed} & The speed at which a player moves without the ball. \\
& & \textbf{Range}: 5.5 to 9.5 m/s \\
\hline
Shoot Speed & \texttt{shoot-speed} & The speed imparted to the ball when a player kicks it. \\
& & \textbf{Range}: 14 to 14.8 m/s \\
\hline
Pass Speed & \texttt{pass-speed} & The speed of the ball when passed. \\
& & \textbf{Range}: 25 to 25.8 m/s \\
\hline
\end{tabular}

\label{table:player_attributes}
\end{table}

\noindent
Based on the information presented in Table~\ref{table:player_attributes}, which outlines the general roles and attributes assigned to players in our simulation, we further categorize the players into specific roles with associated behaviors. To provide a detailed configuration of player roles and team distributions, Table~\ref{t1pr} lists the combinations of defensive or attacking roles with team-oriented or selfish characteristics, along with their respective team assignments.

\begin{table}[H]
\centering
\caption{Player role, defense/attack, team player/selfish, and team.}
\begin{tabular}{|c|c|c|c|}
\hline
Player Role          & Defense/Attack & Team Player/Selfish & Team \\ \hline
defense-team-blue    & Defense        & Team Player         & Blue \\ \hline
defense-selfish-blue & Defense        & Selfish             & Blue \\ \hline
attack-team-blue     & Attack         & Team Player         & Blue \\ \hline
attack-selfish-blue  & Attack         & Selfish             & Blue \\ \hline
defense-team-red     & Defense        & Team Player         & Red  \\ \hline
defense-selfish-red  & Defense        & Selfish             & Red  \\ \hline
attack-team-red      & Attack         & Team Player         & Red  \\ \hline
attack-selfish-red   & Attack         & Selfish             & Red  \\ \hline
\end{tabular}

\label{t1pr}
\end{table}

\noindent
Building upon these roles, we define the behavioral tendencies of each player type to simulate realistic decision-making processes. Table~\ref{t2ppp} presents the action probabilities assigned to each player role, specifying the likelihood of shooting, dribbling, or passing the ball, particularly when far from the goal. These probabilities are integral to reflecting the players' roles and personal tendencies within the game dynamics.

\begin{table}[H]
\centering
\caption{Player shooting, dribbling, and passing probabilities.}
\begin{tabular}{|c|c|c|c|}
\hline
Player Role          & \begin{tabular}[c]{@{}c@{}}Shoot \\ Probability\end{tabular} & \begin{tabular}[c]{@{}c@{}}Dribble \\ Probability\end{tabular} &  \begin{tabular}[c]{@{}c@{}}Pass Probability \\ (Far from the Goal)\end{tabular} \\ \hline
defense-team-blue    & 10\%              & 40\%                & 20\%                                 \\ \hline
defense-selfish-blue & 30\%              & 40\%                & 5\%                                  \\ \hline
attack-team-blue     & 10\%              & 40\%                & 20\%                                 \\ \hline
attack-selfish-blue  & 30\%              & 40\%                & 5\%                                  \\ \hline
defense-team-red     & 10\%              & 40\%                & 20\%                                 \\ \hline
defense-selfish-red  & 30\%              & 40\%                & 5\%                                  \\ \hline
attack-team-red      & 10\%              & 40\%                & 20\%                                 \\ \hline
attack-selfish-red   & 30\%              & 40\%                & 5\%                                  \\ \hline
\end{tabular}

\label{t2ppp}
\end{table}

These tables collectively provide a comprehensive framework for player behavior in the simulation. By defining specific roles and associated probabilities, we ensure that each player's actions are consistent with their attributes and the overall team strategy. This layered approach allows for nuanced interactions within the game, contributing to a more realistic and dynamic simulation environment. For instance, some players are identified for their defensive skills with a strong inclination toward teamwork, while others are noted for their offensive capabilities but prefer to act solo. This setup allows for the customization of team dynamics and strategies, enabling a detailed analysis of how individual behaviors and team interactions influence the game’s outcomes and the mechanics of collisions within the rugby context. This approach enhances the understanding of strategic plays and player positioning, crucial for studying sports collisions in real scenarios.

\subsubsection{ \bf {Ball Mechanics}}
The ball is initialized at a predefined position and follows specific dynamics based on player interactions. The ball follows the player who is currently holding it. When a player shoots or passes, the ball moves towards a target with a speed based on the player's shoot-speed or pass-speed.The ball's movement is updated each tick, considering its flying status and target coordinates.

\subsubsection{ \bf {Game Mechanics and Player Interaction Dynamics}}
\begin{figure}[H]
\centering
\includegraphics[width=2.8 in]{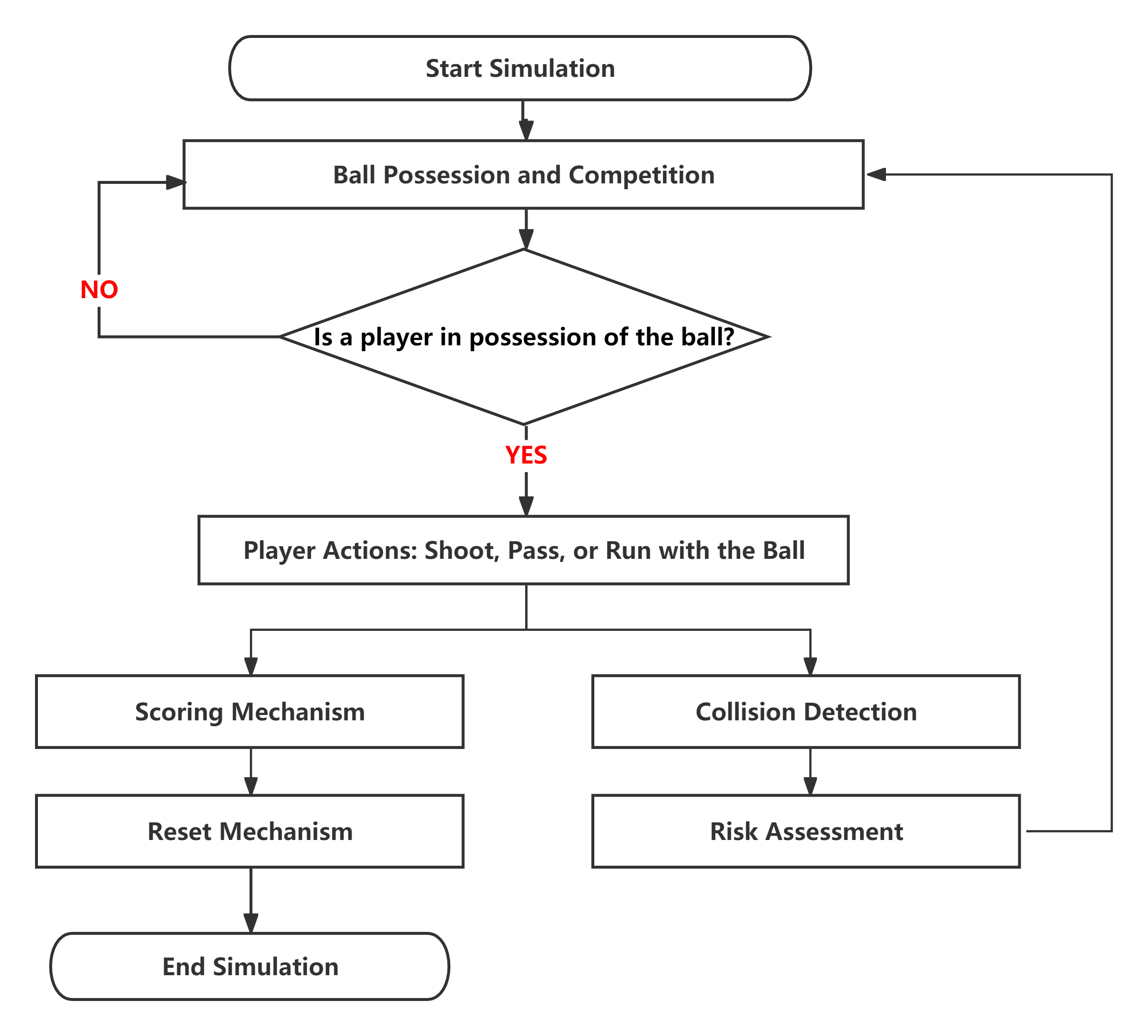}
\caption{Flowchart of Game Mechanics and Collision Detection with Risk Assessment}
\label{fig:game_flowchart}
\end{figure}

Figure~\ref{fig:game_flowchart} illustrates the core mechanics of our rugby simulation model, highlighting the integration of game dynamics with collision detection and risk assessment processes. The flowchart provides a visual representation of the sequential steps and decision points that govern player interactions, ball possession, and scoring within the simulation environment. By delineating these processes, we aim to clarify how individual player attributes and actions contribute to the overall game flow and how these, in turn, influence drone behaviors in the Drone Model.

At the commencement of the simulation, the Ball Possession and Competition mechanism is activated, where players vie for control of the ball. A critical decision point assesses whether a player has gained possession. If possession is established, the simulation progresses to Player Actions, where the player decides to shoot, pass, or run with the ball based on their attributes and proximity to the goal. This decision-making process is essential for simulating realistic player behaviors and strategic gameplay.

Subsequent to player actions, the Scoring Mechanism evaluates the outcome, awarding points when appropriate—such as when a player carrying the ball enters the try zone or when the ball reaches the goal area without an owner. Following a scoring event, the Reset Mechanism reinitializes player positions and ball status, preparing the simulation for the next phase of play. This cyclical process ensures continuous gameplay and allows for the analysis of multiple game scenarios within a single simulation run.

Parallel to the main game mechanics, the flowchart incorporates Collision Detection and Risk Assessment processes. After player actions, collision detection algorithms identify any physical interactions between players within a certain distance, impacting subsequent movement decisions. The risk assessment then marks players as high-risk during ball competitions, which influences drone behaviors in the Drone Model. This integration ensures that drones respond dynamically to the evolving game state, enhancing the realism and complexity of the simulation.

The flowchart presented in Figure~\ref{fig:game_flowchart} encapsulates the interplay between game mechanics and the supplementary processes of collision detection and risk assessment within our rugby simulation. By integrating these components, we achieve a comprehensive model that not only simulates player behaviors and game outcomes but also facilitates the dynamic interaction between players and drones. The decision points and feedback loops highlight the simulation's ability to adapt to changing conditions, reflecting the unpredictable nature of real-world rugby matches. The detailed representation of these processes sets the foundation for the subsequent sections, where we delve deeper into the Drone Model and its integration with the rugby simulation.

\subsection{ \bf {The Drone Model Algorithms}}
In this section, we introduce the algorithms that govern the behavior of drones within our simulation environment. Each algorithm corresponds to a specific operational mode, designed to emulate different surveillance and tracking strategies during a rugby match. The modes include \texttt{\textbf{Fixed Mode}}, \texttt{\textbf{Follow-Ball Mode}}, \texttt{\textbf{Follow-Players Mode}}, \texttt{\textbf{Density-Based Mode}}, \texttt{\textbf{Repulsive Mode}}, and \texttt{\textbf{Random Mode}}. 

\ 
Additionally, the parameters such as the number of drones, their speed, and detection radius are adjustable to tailor the drone fleet's operations to the specific requirements of each simulation scenario, enhancing the fidelity and utility of the captured data for subsequent analysis.

\
\

\begin{table}[h]
\centering
\caption{Descriptions of All Presented Drone Behaviour Algorithms}
\begin{tabular}{|>{\centering\arraybackslash}p{1.5cm}| >{\centering\arraybackslash}p{2 cm} |p{5 cm}|}
\hline
\textbf{Algorithm \newline Number} & \textbf{Mode Name} & \textbf{Description} \\
\hline
1 & \texttt{\textbf{Fixed}} & Drones remain stationary at predefined coordinates. \\
\hline
2 & \texttt{\textbf{Follow-Ball}} & Drones form a formation around the ball,  maintaining equal spacing while following its movement. \\
\hline
3 & \texttt{\textbf{Follow-Players}} & Drones track high-risk players identified during ball competitions. \\
\hline
4 & \texttt{\textbf{Density-Based}} & Drones allocate themselves around player groups based on density levels, focusing on areas with higher player concentration. \\
\hline
5 & \texttt{\textbf{Repulsive}} & Drones follow the ball while avoiding collisions through repulsive forces from other drones. \\
\hline
6 & \texttt{\textbf{Random}} & Drones move randomly within the field boundaries, incorporating collision avoidance mechanisms. \\
\hline
\end{tabular}

\label{table:drone_algorithms}
\end{table}

The descriptions of these operational modes are summarized in Table \ref{table:drone_algorithms}, which outlines the mode name, algorithm number, and a brief description of each mode. These algorithms provide detailed procedural steps for drone positioning and movement, ensuring that drones interact with players and the ball in a manner consistent with their designated roles. By formalizing these algorithms, we enable reproducible and scalable simulation of drone behaviours for analysis and optimization in various scenarios. Note that we make the following assumptions: (1) Drones can autonomously ﬂy to the next location without collision, (2) Camera orientation is ﬁxed on the drones, and (3) drones operate at a constant height.

Below, we provide detailed algorithmic steps for each drone mode, accompanied by brief introductions that explain the purpose and functionality of each mode.

\subsubsection{ \texttt{\textbf {Fixed Mode Algorithm}}}
In \texttt{Fixed Mode}, drones are strategically deployed at fixed positions to maximize coverage in areas with high collision frequencies. This mode leverages collision data from prior simulations to determine optimal drone placement, ensuring enhanced surveillance in regions where it is most needed. While prior work, such as the multi-camera tracking system for football games by by Takahashi et al. (2018)\cite{fixedref}, achieved significant improvements in ball tracking and real-time analytics for live broadcasts, it faced notable challenges. Their system, reliant on consumer-grade HD cameras and integration algorithms, struggled with:
1. Occlusions: Ball visibility was compromised in crowded player regions or during long-term obstructions.
2. Limited Precision: With an average error of 5.3 meters, the system was unsuitable for applications requiring finer spatial resolution, such as offside detection or goal-line tracking.
3. Environmental Sensitivity: Varying lighting conditions, such as shadows and artificial illumination, impacted robustness.
4. Deployment Complexity: The requirement for precise camera calibration and a dense network of devices restricted the system's scalability and cost-effectiveness.

Our proposed method overcomes these limitations by using drones equipped with a flexible deployment framework. Unlike fixed camera setups, drones can reposition dynamically, adjust their coverage zones, and maintain visibility even in occluded or dynamic scenarios. This adaptability makes drones particularly advantageous in environments with non-uniform collision distributions or unexpected changes, such as player density shifts or adverse weather conditions.

The process involves analyzing collision coordinates from the simulation, represented as \( C = \{(x_c, y_c)\} \), where \( (x_c, y_c) \) denotes the coordinates of each collision point. These coordinates are quantized to integer grid points:

\begin{equation}
\label{eq:quantize}
(x_q, y_q) = (\lfloor x_c \rfloor, \lfloor y_c \rfloor),
\end{equation}

where \( (x_q, y_q) \) is the quantized coordinate. A collision frequency map \( F(x_q, y_q) \) is then generated, which counts the number of collisions at each grid point:

\begin{equation}
\label{eq:collision_map}
F(x_q, y_q) = \sum_{(x_c, y_c) \in C} \delta(x_c, x_q) \cdot \delta(y_c, y_q),
\end{equation}

where \( \delta(a, b) \) is the Kronecker delta, defined as:

\begin{equation}
\label{eq:kronecker_delta}
\delta(a, b) =
\begin{cases}
1, & \text{if } a = b, \\
0, & \text{otherwise}.
\end{cases}
\end{equation}

Next, for each grid point \( (x_i, y_i) \), the coverage \( S_i \) is computed as the number of uncovered collision points within a radius \( r \):

\begin{equation}
\label{eq:coverage}
S_i = \left| \left\{ (x_c, y_c) \in C_{\text{uncovered}} \mid \sqrt{(x_c - x_i)^2 + (y_c - y_i)^2} \leq r \right\} \right|,
\end{equation}

where \( |\cdot| \) denotes the cardinality of the set, i.e., the number of elements in the set of uncovered collision points that are within radius \( r \) from the grid point \( (x_i, y_i) \).

The algorithm iteratively selects the grid point \( (x_{\text{max}}, y_{\text{max}}) \) with the maximum coverage \( S_{\text{max}} \), and places a drone at that position. The selected position is then added to the drone position list \( D \), and all collision points within radius \( r \) are marked as covered.

\begin{algorithm}[h] \caption{\texttt{Fixed Mode} Drone Positioning Based on Collision Data} \begin{algorithmic}[1] 
\STATE \textbf{Input:} Collision data set $C = \{(x_c, y_c)\}$, number of drones $N$, drone coverage radius $r$
\STATE \textbf{Initialize:} Mark all collision points in $C$ as uncovered. \STATE Quantize collision positions to integer coordinates to form a grid  \( (x_q, y_q) \)(Eqn. \eqref{eq:quantize}). 
\STATE Accumulate collision counts at each grid coordinate, resulting in a collision frequency map \( F(x_q, y_q) \)(Eqn. \eqref{eq:collision_map}). 
\STATE Initialize drone position list $D = \{\}$.
\WHILE{there are uncovered collision points \textbf{and} $|D| < N$} \FOR{each grid coordinate $(x_i, y_i)$ in $F$} 
\STATE Compute coverage $S_i$ as the number of uncovered collision counts within radius $r$ centered at $(x_i, y_i)$ (Eqn. \eqref{eq:coverage}). \ENDFOR 
\STATE Select coordinate $(x_{\text{max}}, y_{\text{max}})$ with maximum coverage $S_{\text{max}}$. 
\STATE Add $(x_{\text{max}}, y_{\text{max}})$ to drone position list $D$. \STATE Mark collision points within radius $r$ of $(x_{\text{max}}, y_{\text{max}})$ as covered. \ENDWHILE 
\STATE \textbf{Output:} Drone positions $D$ 
\end{algorithmic} \end{algorithm}

\begin{figure}[h]
\centering
\includegraphics[width=2.5in]{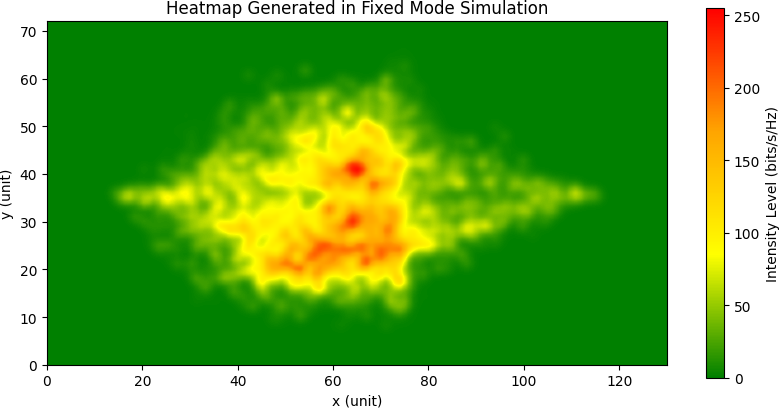}
\caption{Generated Heatmap Based on Collision Data}
\label{fixedheatmap}
\end{figure}

The algorithm operates by quantizing collision data to create a collision frequency heatmap. Figure \ref{fixedheatmap} illustrates the generated heatmap, highlighting areas on the field with the highest frequency of collisions.

\begin{figure}[h]
\centering
\includegraphics[width=2.5in]{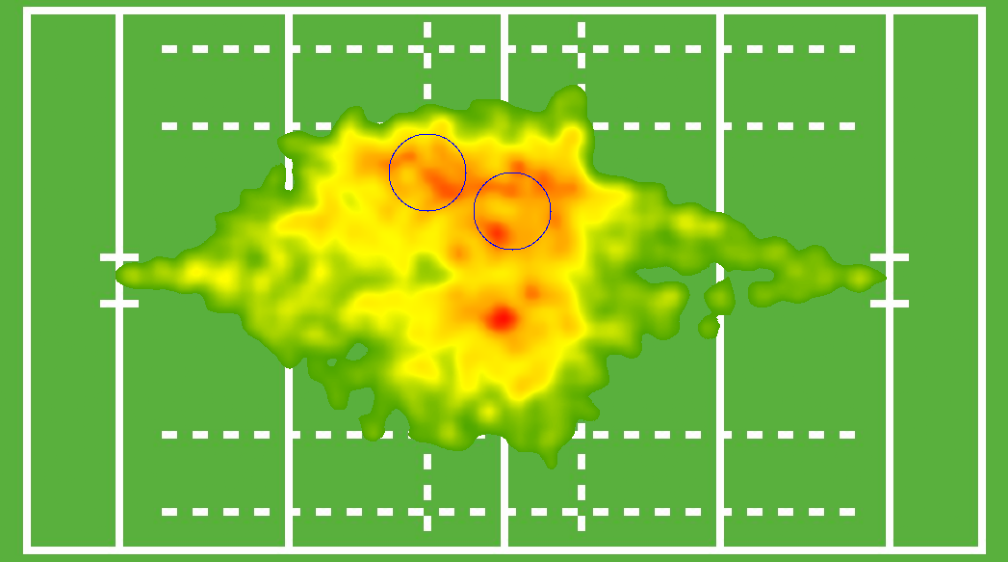}
\caption{Simulation Results of Fixed UAV Positions (Fixed Mode) with 2 Drones in the radius of 5}
\label{Fixedoutput1}
\end{figure}

Using the heatmap, the algorithm identifies grid points that cover the maximum number of collisions within the drone's coverage radius. Drones are then positioned at these optimal locations to ensure maximum surveillance coverage. Figure \ref{Fixedoutput1} shows the layout of drones in our simulation, where 2 drones are strategically placed with a radius of 5 units.

\subsubsection{ \texttt{\textbf {Follow-Ball Mode Algorithm}}}

The \texttt{Follow-Ball} Mode is inspired by current practices in live sports broadcasting, where cameras closely track the ball to capture exciting moments during the game. Similarly, in our model, drones naturally follow the rugby ball, maintaining a tight formation at a fixed radius \( R \) (the drone formation radius) around the ball. This ensures continuous and focused surveillance of the ball's immediate vicinity, allowing for real-time monitoring of crucial game events.

\begin{algorithm}[h]
\caption{\texttt{Follow-Ball Mode} Drone Movement}
\begin{algorithmic}[1]
\STATE \textbf{Input:} Number of drones \( N \), formation radius \( R \) (\texttt{radius-of-drones})
\FOR{each simulation tick}
    \STATE Obtain current ball position \( (x_b, y_b) \).
    \STATE Calculate angle step \( \theta_{\text{step}} = \frac{2\pi}{N} \).
    \FOR{each drone \( d_i \), \( i = 1 \) to \( N \)}
        \STATE Compute angle \( \theta_i = \theta_{\text{step}} \times (i - 1) \).
        \STATE Update drone position:
        \STATE \quad \( x_i = x_b + R \times \cos(\theta_i) \),
        \STATE \quad \( y_i = y_b + R \times \sin(\theta_i) \).
    \ENDFOR
\ENDFOR
\end{algorithmic}
\end{algorithm}

This algorithm positions drones in a circular formation around the ball, ensuring equal spacing and synchronized movement as the ball moves. By maintaining a constant distance \( R \) from the ball, drones provide comprehensive coverage of the area where pivotal actions are most likely to occur. This approach leverages the dynamic nature of the game, allowing drones to adaptively reposition themselves in response to the ball's movement while maintaining formation integrity.

\subsubsection{ \texttt{\textbf {Repulsive Mode Algorithm}}}
The \texttt{Repulsive} Mode is designed to address the limitations of the {\texttt{Follow-Ball}} Mode, where drones following the same target may inadvertently collide or overlap due to converging paths. By integrating collision avoidance, drones can maintain optimal positioning around the ball without interfering with each other's flight paths. This mode assumes that drones can detect neighboring drones within a certain radius and adjust their movements accordingly to prevent collisions.

{ \bf {Collision Avoidance Mechanism}}

While following the ball, drones need to avoid close proximity with other drones to prevent overlap and potential collisions. The collision avoidance is achieved through the following steps:

1. Detection of Neighboring Drones: Each drone identifies other drones within a specified detection radius, typically set to twice the operational radius of a drone (\(2r\)), where \(r\) is the drone's coverage radius.

2. Computing Repulsive Movement: If neighbouring drones are detected within this radius, the drone computes the centre of mass of these neighbours. It then adjusts its heading to move away from this centre of mass, effectively increasing separation.

3. Randomized Movement Distance: The drone moves a random distance proportional to how close it is to the neighbors, adding randomness to prevent synchronized movements that could lead to new collision courses.

The rationale behind using the centre of mass is to provide a general direction for avoidance, simplifying calculations and ensuring efficient dispersal of drones when they are too close.

{ \bf {\texttt{Repulsive Mode} Algorithm Description}}

The algorithm operates in two main phases during each simulation tick: following the ball and collision avoidance.

\begin{algorithm}[h]
\caption{\texttt{Repulsive Mode} Drone Movement with Collision Avoidance}
\begin{algorithmic}[1]
\STATE \textbf{Input:} Number of drones \( N \), drone radius \( r \) (\texttt{radius-of-drones}), maximum movement distance \( d_{\text{max}} \), simulation time step \( \Delta t \)
\FOR{each simulation tick}
    \STATE \textbf{Phase 1: Follow the Ball}
    \FOR{each drone \( d_i \)}
        \STATE Obtain current position \( (x_i, y_i) \)
        \STATE Obtain ball position \( (x_b, y_b) \)
        \STATE Move towards the ball using the function \texttt{FollowBall}:
        \STATE \quad \( (x_i, y_i) \leftarrow \text{FollowBall}(x_i, y_i, x_b, y_b) \)
    \ENDFOR
    \STATE \textbf{Phase 2: Collision Avoidance}
    \FOR{each drone \( d_i \)}
        \STATE Identify neighboring drones \( D_{\text{near}} \) within distance \( 2r \)
        \IF{\( D_{\text{near}} \) is not empty}
            \STATE Compute center of mass of neighbors:
            \STATE \quad \( x_{\text{mean}} = \dfrac{1}{|D_{\text{near}}|} \sum_{d_j \in D_{\text{near}}} x_j \)
            \STATE \quad \( y_{\text{mean}} = \dfrac{1}{|D_{\text{near}}|} \sum_{d_j \in D_{\text{near}}} y_j \)
            \STATE Compute distance to center of mass:
            \STATE \quad \( d_{\text{mean}} = \text{distance}((x_i, y_i), (x_{\text{mean}}, y_{\text{mean}})) \)
            \IF{\( d_{\text{mean}} < 2r \)}
                \STATE Compute heading away from center of mass:
                \STATE \quad \( \theta_i = \text{atan2}(y_i - y_{\text{mean}},\, x_i - x_{\text{mean}}) \)
                \STATE Compute random movement distance:
                \STATE \quad \( d_{\text{move}} = \text{random}(0,\, 2r - d_{\text{mean}}) \)
                \STATE Update position:
                \STATE \quad \( x_i \leftarrow x_i + d_{\text{move}} \times \cos(\theta_i) \)
                \STATE \quad \( y_i \leftarrow y_i + d_{\text{move}} \times \sin(\theta_i) \)
            \ENDIF
        \ENDIF
    \ENDFOR
\ENDFOR
\end{algorithmic}
\end{algorithm}

The \texttt{Repulsive Mode} algorithm is designed with computational efficiency in mind, leveraging efficient spatial search techniques to identify neighboring drones, which ensures real-time performance even when multiple drones are in operation. Drones continuously monitor their positions relative to the field boundaries, adjusting movements as necessary to remain within the operational area and maintain boundary compliance. The algorithm supports dynamic adaptation, allowing drones to adjust their paths in response to the movements of both the ball and other drones, thereby maintaining focus on the target while ensuring safe separation. Introducing randomness into the movement distance is essential to prevent deterministic patterns that could lead to synchronization issues or new collision courses; this stochastic element enhances the realism of drone behavior within the simulation.

The \texttt{Repulsive Mode} Algorithm effectively enhances the \texttt{Follow-Ball Mode} by incorporating a collision avoidance mechanism. By detecting neighboring drones within a specified radius and adjusting movements away from the center of mass of nearby drones, the algorithm ensures safe separation while maintaining focus on the ball. The use of randomized movement distances prevents synchronization issues, and the continuous boundary checks keep drones within the operational field. This mode offers significant advantages in scenarios where multiple drones are required to follow the same target without overlapping, providing a balance between coverage efficiency and operational safety.

\subsubsection{ \texttt{\textbf {\texttt{Follow-Players Mode} Algorithm}}}
The \texttt{Follow-Players Mode} is designed for drones to track high-risk players identified during ball competitions. This mode enhances surveillance by focusing on players who are most likely to impact the game's outcome during critical moments. The algorithm distinguishes between high-risk and low-risk players based on their proximity to the ball and their strategic positioning relative to teammates.

The identification of high-risk players involves the following logic:

\begin{figure}[h]
\centering
\includegraphics[width=2.65 in]{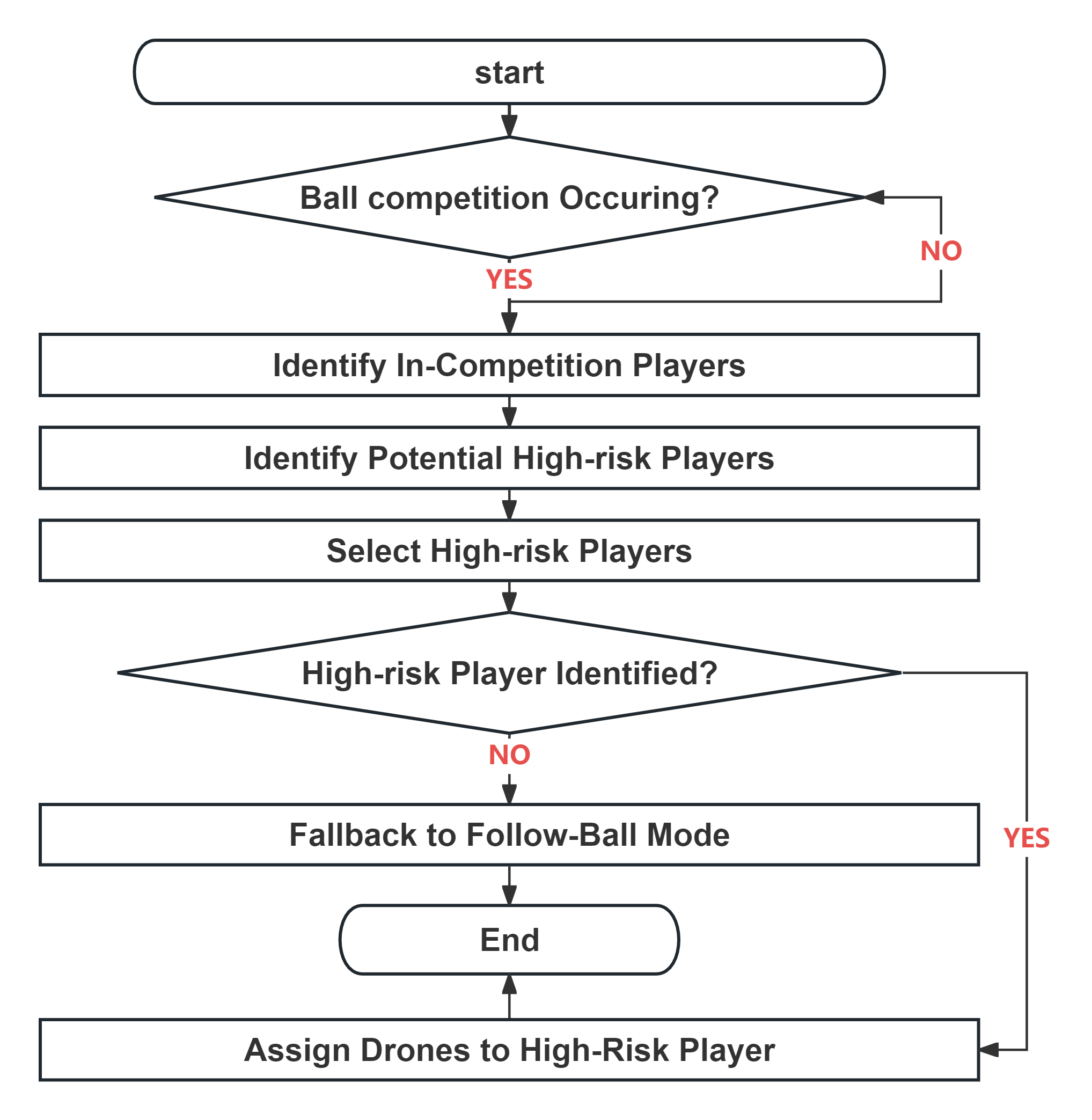}
\caption{Flowchart for Identifying High-Risk Players and Drone Assignment}
\label{fig:high_risk_flowchart}
\end{figure}

The \texttt{Follow-Players Mode} algorithm, as illustrated in Fig. \ref{fig:high_risk_flowchart}, begins by detecting whether a ball competition is occurring, identifying situations where players are actively contesting possession of the ball. Players within close proximity to the ball (e.g., within 3 units) are classified as "in-competition players." For each of these players, the algorithm searches for the nearest teammate located outside the immediate competition area but within a specified range (e.g., between 3 and 15 units from the ball). It then calculates the cumulative distance from the ball to the in-competition player and from that player to their nearest teammate. The player with the minimum cumulative distance is designated as the high-risk player, as they are in a strategic position to receive the ball or significantly influence the game's outcome. Once identified, drones are assigned to form a formation around the high-risk player. If no high-risk players are identified, the drones default to following the ball, ensuring continuous monitoring and adaptability to the game's dynamics.

The core idea is to have drones naturally follow high-risk players by maintaining a formation within a radius R (the drone formation radius) around the identified player. This approach ensures that drones provide focused surveillance on players who are likely to make pivotal moves during the game.

\begin{algorithm}[h]
\caption{\texttt{Follow-Players Mode} Drone Movement}
\begin{algorithmic}[1]
\STATE \textbf{Input:} Number of drones $N$, formation radius $R$
\FOR{each simulation tick}
    \IF{Ball competition is occurring}
        \STATE Identify in-competition players $P_{\text{in}}$ within distance $d_{\text{in}}$ (e.g., 3 units) from the ball.
        \STATE Identify other players $P_{\text{out}}$ within distance $d_{\text{out}}$ (e.g., between 3 and 15 units) from the ball.
        \STATE Initialize minimum cumulative distance $d_{\text{min}} \gets \infty$.
        \STATE Initialize high-risk player $p_{\text{hr}} \gets \text{null}$.
        \FOR{each player $p_i$ in $P_{\text{in}}$}
            \STATE Find nearest teammate $p_j$ in $P_{\text{out}}$ such that $p_j.\text{team} = p_i.\text{team}$.
            \IF{$p_j$ exists}
                \STATE Compute cumulative distance $d = \text{distance}(p_i, \text{ball}) + \text{distance}(p_i, p_j)$.
                \IF{$d < d_{\text{min}}$}
                    \STATE $d_{\text{min}} \gets d$.
                    \STATE $p_{\text{hr}} \gets p_i$.
                \ENDIF
            \ENDIF
        \ENDFOR
        \IF{$p_{\text{hr}}$ is not \text{null}}
            \STATE \textbf{Assign drones to high-risk player} $p_{\text{hr}}$:
            \STATE Calculate angle step $\theta_{\text{step}} = \frac{2\pi}{N}$.
            \FOR{each drone $d_k$, $k = 1$ to $N$}
                \STATE Compute angle $\theta_k = \theta_{\text{step}} \times (k - 1)$.
                \STATE Update drone position:
                \STATE \quad $x_k = x_{p_{\text{hr}}} + R \times \cos(\theta_k)$,
                \STATE \quad $y_k = y_{p_{\text{hr}}} + R \times \sin(\theta_k)$.
            \ENDFOR
        \ELSE
            \STATE \textbf{Fallback to Follow-Ball Mode}:
            \STATE Execute Algorithm 2 with all drones.
        \ENDIF
    \ELSE
        \STATE \textbf{Fallback to Follow-Ball Mode}:
        \STATE Execute Algorithm 2 with all drones.
    \ENDIF
\ENDFOR
\end{algorithmic}
\end{algorithm}

\textbf{\texttt{Follow-Players Mode} Algorithm Description}: 

The proposed algorithm dynamically adjusts drone movements based on the game's state to enhance surveillance and data collection. Initially, the algorithm detects whether a ball competition is in progress. If active competition is identified, players are classified into two sets: \( P_{\text{in}} \), representing those within a distance \( d_{\text{in}} \) (e.g., 3 units) from the ball, and \( P_{\text{out}} \), representing nearby players within \( d_{\text{out}} \) (e.g., 3–15 units) but not directly involved in the competition.

For each player \( p_i \in P_{\text{in}} \), the algorithm locates the nearest teammate \( p_j \in P_{\text{out}} \) who belongs to the same team. It then calculates a cumulative distance \( d \), defined as the sum of the distance from \( p_i \) to the ball and from \( p_i \) to \( p_j \). The player \( p_{\text{hr}} \) with the minimum \( d \) is deemed the high-risk player, having the greatest potential to impact the game.

Drones are assigned to form a circular formation around \( p_{\text{hr}} \), with positions calculated based on the number of drones \( N \) and formation radius \( R \). If no high-risk player is identified, drones revert to a default mode that maintains surveillance around the ball. This adaptive mechanism ensures that drones focus on key players during critical moments, optimizing their coverage and strategic value.

\newpage
\begin{table}[h]
    \centering
    \caption{Explanation of Symbols and Variables}
    \setlength{\tabcolsep}{0.5pt} 
    \renewcommand{\arraystretch}{0.7} 
    \footnotesize 
    \begin{tabularx}{\linewidth}{lXlX}
        \hline
        \textbf{Symbol} & \textbf{Explanation} & \textbf{Symbol} & \textbf{Explanation} \\
        \hline
        $N$ & Total number of drones & $R$ & Formation radius around the high-risk player \\
        $P_{\text{in}}$ & Set of in-competition players & $P_{\text{out}}$ & Set of other players near the ball \\
        $d_{\text{in}}$ & Distance threshold for $P_{\text{in}}$ & $d_{\text{out}}$ & Distance threshold for $P_{\text{out}}$ \\
        $d_{\text{min}}$ & Minimum cumulative distance & $p_{\text{hr}}$ & High-risk player \\
        $p_{i}$ & A player in $P_{\text{in}}$ & $p_{j}$ & Nearest teammate in $P_{\text{out}}$ \\
        $\theta_{\text{step}}$ & Angle step for drone placement & $\theta_{k}$ & Angle for drone $d_{k}$ \\
        $x_{k}, y_{k}$ & Coordinates of drone $d_{k}$ & $X_{P_{\text{hr}}}, Y_{P_{\text{hr}}}$ & Coordinates of $P_{\text{hr}}$ \\
        \hline
    \end{tabularx}
    \label{table:symbols}
\end{table}

\subsubsection{ \texttt{\textbf {Density-Based Mode Algorithm}}}
In the \texttt{Density-Based Mode}, drones dynamically allocate themselves around regions of high player density to optimize surveillance. The algorithm identifies up to four density centers based on player clustering and assigns drones to these centers proportionally, with more drones allocated to regions of higher density.

\begin{algorithm}[h]
\caption{\texttt{Density-Based Mode} Drone Movement}
\begin{algorithmic}[1]
\STATE \textbf{Input:} Number of drones $N$, radius $r$, player positions set $P$.
\FOR{each simulation tick}
    \STATE Cluster players into groups $C = \{C_1, C_2, ..., C_K\}$ based on proximity.
    \STATE Compute density $\rho_j = |C_j|$ for each cluster $C_j$.
    \STATE Sort clusters by density in descending order.
    \STATE Allocate drones to clusters proportionally to density:
    \STATE \quad $D_j = \left\lfloor N \times \dfrac{\rho_j}{\sum_{k=1}^{K} \rho_k} \right\rfloor$.
    \STATE $k \leftarrow 1$ \COMMENT{Drone index}
    \FOR{each cluster $C_j$}
        \STATE Compute cluster centroid $(x_{C_j}, y_{C_j})$.
        \STATE Calculate angle step $\theta_{\text{step}} = \dfrac{2\pi}{D_j}$.
        \FOR{$i = 1$ to $D_j$}
            \STATE Compute angle $\theta_i = \theta_{\text{step}} \times (i - 1)$.
            \STATE Update drone position:
            \STATE \quad $x_k = x_{C_j} + r \times \cos(\theta_i)$,
            \STATE \quad $y_k = y_{C_j} + r \times \sin(\theta_i)$.
            \STATE $k \leftarrow k + 1$.
        \ENDFOR
    \ENDFOR
\ENDFOR
\end{algorithmic}
\end{algorithm}
In this \texttt{Density-Based Mode} Algorithm, initially, all players are marked as non-density centres, and then an empty list of excluded players is created to keep track of those already associated with density centres. The density level index \( L \) starts at 0, representing the highest density level.

The algorithm searches for up to four density centres. In each iteration, it scans all players not yet excluded and counts the number of neighbouring players within the density detection radius for each. The player with the highest neighbour count is selected as the density center for that level. All players within this radius are added to the excluded list to prevent overlapping density centres, ensuring they are spread out across the field.

Drones are assigned to density levels using a hierarchical allocation strategy. For levels 0 to 2, the number of drones assigned is half of the remaining drones at each subsequent level. Level 0 receives half of the available drones, level 1 receives half of the remaining drones, and level 2 follows the same pattern. Level 3, the last level, receives all remaining drones. This approach prioritizes higher-density areas by assigning them more drones. Each drone is assigned a `followLevel` corresponding to the density level and a `followIdx` to determine its position around the density centre.

Drones assigned to a density center are positioned in a circular formation around the centre. The angle between each drone is calculated to ensure they are evenly spaced. Using trigonometric functions and the specified radius \( r \), target positions are computed. Drones move towards these positions and maintain formation as the density centers (players) move.

Therefore, through positioning drones in evenly spaced circular formations around density centers, the algorithm enhances surveillance coverage while minimizing the risk of drone collisions. Key parameters, such as the density detection radius and the maximum number of density levels, can be adjusted to meet specific surveillance requirements, making the algorithm flexible and scalable for various scenarios.

\subsubsection{ \texttt{\textbf {Random Mode Algorithm}}}
The \texttt{Random Mode} is designed to emulate unpredictable drone movements within the operational field. Unlike other modes that have specific targets or formations, drones in \texttt{Random Mode} move towards randomly selected positions while avoiding collisions with other drones and staying within field boundaries. This randomness provides a robust testing environment for collision avoidance mechanisms and helps in assessing the drones' ability to navigate autonomously without predefined paths.

Before presenting the algorithm, we outline the key components of the movement strategy and collision avoidance mechanisms:

Random Target Selection: Each drone selects a random target location within the field boundaries that is unoccupied and maintains a safe distance from other drones.

Path Adjustment: Drones compute the direction vector towards their target and move accordingly, adding random perturbations to simulate natural movement.

Collision Avoidance: While moving, drones continuously check for nearby drones within a specified safe distance. If another drone is detected within this range, the drone adjusts its movement to prevent collisions.

Boundary Compliance: Drones ensure they remain within the operational field boundaries by adjusting their positions if a movement would result in exiting the area.

\begin{algorithm}[h]
\caption{\texttt{Random Mode} Drone Movement}
\begin{algorithmic}[1]
\STATE \textbf{Input:} Number of drones $N$, maximum speed $v_{\text{max}}$, field boundaries.
\FOR{each simulation tick}
    \FOR{each drone $d_i$}
        \IF{no target assigned or target reached}
            \STATE Randomly select unoccupied target position $(x_{\text{target}_i}, y_{\text{target}_i})$ within field boundaries.
        \ENDIF
        \STATE \textbf{Compute Direction Vector:}
        \STATE $\mathbf{D}_i = \dfrac{(x_{\text{target}_i} - x_i,\, y_{\text{target}_i} - y_i)}{\| (x_{\text{target}_i} - x_i,\, y_{\text{target}_i} - y_i) \|}$.
        \STATE Add random perturbation to $\mathbf{D}_i$.
        \STATE \textbf{Check for Collisions:} $\mathbf{R}_i = \mathbf{0}$.
        \FOR{each drone $d_j$, $j \ne i$}
            \STATE Compute distance $d_{ij} = \| (x_j - x_i,\, y_j - y_i) \|$.
            \IF{$d_{ij} < d_{\text{safe}}$}
                \STATE Compute repulsion $\mathbf{r}_{ij} = \dfrac{(x_i - x_j,\, y_i - y_j)}{d_{ij}^3}$.
                \STATE $\mathbf{R}_i = \mathbf{R}_i + \mathbf{r}_{ij}$.
            \ENDIF
        \ENDFOR
        \STATE \textbf{Update Velocity:} $\mathbf{V}_i = v_{\text{max}} \times \mathbf{D}_i + \mathbf{R}_i$.
        \STATE \textbf{Update Position:} $(x_i, y_i) = (x_i, y_i) + \mathbf{V}_i \times \Delta t$.
    \ENDFOR
\ENDFOR
\end{algorithmic}
\end{algorithm}

The \texttt{Random Mode} Algorithm is designed with computational efficiency in mind, efficiently selecting target positions and computing repulsion vectors to ensure real-time performance. By limiting collision checks to drones within the safe distance \( d_{\text{safe}} \), the algorithm minimizes computational overhead, enabling scalability with multiple drones. Drones select unoccupied target positions that maintain a minimum safe distance from other drones, reducing the likelihood of immediate collisions upon arrival at the target location. Introducing randomness to the direction vector simulates natural movement patterns and prevents drones from following predictable paths that could lead to synchronization issues or collision courses. Furthermore, drones continuously assess their surroundings and adjust their movements to avoid collisions, demonstrating dynamic adaptation and autonomous navigation capabilities. Boundary management is also an integral part of the algorithm; drones ensure they remain within the field boundaries by checking their positions after each movement and making necessary adjustments to stay within the operational area.

\subsection{ \bf {Drone Power Consumption Model}}

The interaction between rugby players and drones forms a complex system with nonlinear behaviors and emergent properties, crucial for understanding sports collisions. To better understand and optimize UAV operations within such scenarios, we establish a comprehensive energy consumption model based on previous foundational research conducted by Thibbotuwawa et al. \cite{2019energy}. This model accurately accounts for various UAV operational states, including hovering, high-speed steady-level flight, and moderate horizontal movement. These states are crucial to realistically simulating UAV physics and their strategic deployment during rugby matches. 

The primary power equations, adapted from \cite{2019energy}, for each operational state are expressed as follows:

Hovering Power (\(P_{hovering}\)):
\begin{equation}
 P_{hovering} = n \left[ \left( \frac{(w \cdot g)^{3/2}}{\sqrt{2 \cdot \rho \cdot A}} \right) \right]   
\tag{4}
\end{equation}
where \( n \) is the efficiency factor, \( w \) is the weight of the drone, \( g \) is the acceleration due to gravity, \( \rho \) is the air density, and \( A \) is the facing area of the UAV.

High-Speed Flight Power (\(P_{high}\)):
\begin{equation}
 P_{high} = n \left[ \frac{C_d}{C_l} \cdot w \cdot v + \frac{w^2}{\rho \cdot b^2 \cdot v} \right] 
\tag{5}
\end{equation}
where \( C_d \) and \( C_l \) are the drag and lift coefficients respectively, \( \rho \) is the drag due to lift, and \( b \) is the width of the UAV.

Moderate Horizontal Movement Power (\(P_{moderate}\)):
\begin{equation}
 P_{moderate} = n \left[ \frac{1}{2} \cdot C_d \cdot A \cdot D \cdot v^3 + \frac{w^2}{D \cdot b^2 \cdot v} \right] 
\tag{6}
\end{equation}
The interaction between rugby players and drones forms a complex system with nonlinear behaviors and emergent properties. To better understand and optimize UAV operations in sports scenarios, we establish a comprehensive energy consumption model that accounts for various flight states. Below focuses on the DJI Air 3 UAV in the EU region, incorporating theoretical foundations and specific drone parameters for practical applications.

{\bf{Moderate Horizontal Movement Power}}

For moderate horizontal flight, the power consumption is given by the aerodynamic lift-drag theory:
\begin{equation}
P_{moderate} = n \left[ \frac{1}{2} C_D A \rho v^3 + \frac{W^2}{\rho b^2 v} \right]
\tag{7}
\end{equation}

Variable Definitions:
- \(P_{moderate}\): Power required for moderate horizontal movement (W),
- \(n\): Efficiency factor,
- \(C_D\): Drag coefficient (\(C_D = 1.1\)),
- \(A\): Rotor-facing area (\(A = 0.032 \, \text{m}^2\)),
- \(\rho\): Air density (\(\rho = 1.225 \, \text{kg/m}^3\) at sea level),
- \(v\): Speed of the UAV (m/s),
- \(W\): Weight of the UAV (\(W = 0.720 \, \text{kg}\)),
- \(b\): Rotor span (\(b = 0.28 \, \text{m}\)).

By substituting the DJI Air 3’s parameters into the equation, the power model is simplified as follows:

1. First Term:
\begin{equation}
   \frac{1}{2} C_D A \rho = 0.5 \cdot 1.1 \cdot 0.032 \cdot 1.225 = 0.02156
\tag{8.1}
\end{equation}

2. Second Term: 
\begin{equation}
   \frac{W^2}{\rho b^2} = \frac{0.720^2}{1.225 \cdot 0.28^2} = 5.4
\tag{8.2}
\end{equation}

Thus, the final expression for \(P_{moderate}\) as a function of speed \(v\) is:

\begin{equation}
P_{moderate} = n \left[ 0.02156 v^3 + \frac{5.4}{v} \right]
\tag{8.3}\label{foeq}
\end{equation}

This equation effectively models the power requirements for horizontal movement under varying speeds.

The total energy consumption \(E\) and flight time \(t\) for the DJI Air 3 are determined using a moderate power model, with a fixed battery capacity of \(E = 62.6 \, \text{Wh}\):

\begin{equation}
t = \frac{E}{P_{moderate}}
\tag{9.1}
\end{equation}

Substituting \(P_{moderate}\):

\begin{equation}
t(v) = \frac{62.6}{n \left[ 0.02156 v^3 + \frac{5.4}{v} \right]}
\tag{9.2}
\end{equation}

This relationship allows for evaluating the operational flight time at different speeds \(v\).

\begin{figure}[h]
\centering
\includegraphics[width=2.5in]{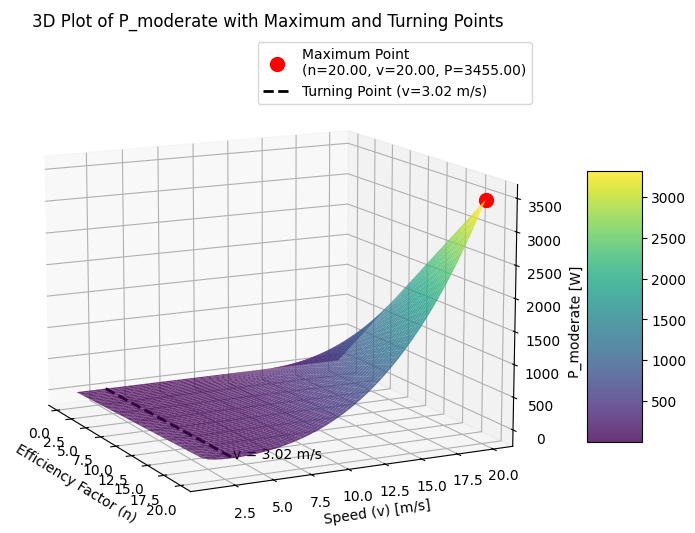}
\caption{Graphical information of Equation \ref{foeq}}
\label{power-s}
\end{figure}

According to the Equation \ref{5.1eq}, \( P_{moderate} \) has a direct linear dependency on the efficiency factor \( n \), implying that increasing \( n \) proportionally elevates the power output. Conversely, the relationship between speed \( v \) and \( P_{moderate} \) is more complex and non-linear. The power output initially experiences a decrease at lower velocities, followed by an inflection point, beyond which it significantly and rapidly increases. However, beyond a specific turning point, which is approximately 3.02 m/s as demonstrated clearly by the plotted surface in Fig.~\ref{power-s}, and then the power output begins to escalate rapidly. Consequently, examining both variables simultaneously reveals a compounded effect: a simultaneous increase in both \( n \) and \( v \) results in a markedly rapid and substantial rise in power output. Conversely, a low efficiency factor can notably suppress the power output, even at higher velocities.

The preceding analysis establishes a baseline for the energy footprint of our solution. In the subsequent discussion, we justify our algorithmic assessment approach, which assumes that an increased number of drones leads to higher energy consumption.
The total energy consumption (\( E \)) of the drones during their operation is calculated by:
\[ E = P_{hovering} \cdot t_{hovering} + P_{fly} \cdot t_{fly} \]
where \( t_{hovering} \) and \( t_{fly} \) are the times spent in hovering and flying states respectively.

Given the total operational time \( T \) as the sum of \( t_{hovering} \) and \( t_{fly} \), we have the relationship:
\[ t_{hovering} + t_{fly} = T \]

This relationship can be normalized by dividing each term by \( T \), yielding:
\[ \frac{t_{hovering}}{T} + \frac{t_{fly}}{T} = 1 \]

Now, consider a scenario where the number of drones increases from \( n \) to \( n+1 \). Assuming the total time \( T \) remains constant, and based on the operational dynamics of drones, the presence of additional drones typically results in an increased need for hovering due to coordination and airspace management. Therefore, as \( n \) increases to \( n+1 \), the hovering time \( t_{hovering} \) is expected to increase while \( t_{fly} \) decreases:
\[ t_{hovering}' > t_{hovering} \]
\[ t_{fly}' < t_{fly} \]

To illustrate, let's express the new times in terms of the changes:
\[ t_{hovering}' = t_{hovering} + \Delta t \]
\[ t_{fly}' = t_{fly} - \Delta t \]

Given that the total time is constant:
\[ t_{hovering}' + t_{fly}' = T \]
\[ (t_{hovering} + \Delta t) + (t_{fly} - \Delta t) = T \]

This simplifies to:
\[ t_{hovering} + t_{fly} = T \]
confirming our initial total time equation.

Now substituting these expressions into the energy consumption formula for \( n+1 \) drones:
\[ E' = P_{hovering} \cdot t_{hovering}' + P_{fly} \cdot t_{fly}' \]
\[ E' = P_{hovering} \cdot (t_{hovering} + \Delta t) + P_{fly} \cdot (t_{fly} - \Delta t) \]

Expanding this expression provides insight into how the changes in \( t_{hovering} \) and \( t_{fly} \) due to an additional drone affect the total energy consumption, reflecting the trade-offs between energy efficiency and {\bf{the number of drones}} deployed. This analysis is crucial for optimizing UAV operations, ensuring that they not only meet the coverage needs but also do so in an energy-efficient manner, balancing the benefits of additional drones against their cost in terms of energy.

\

\section{Results}

In this section, we comprehensively analyze the collision detection accuracy of six UAV tracking strategies: \texttt{\textbf{Density-based, Follow-ball, Fixed (Heat Map), Follow-players, Random, and Repulsive}}.

\subsection{Comprehensive Analysis of UAV Collision Detection Accuracy under Varying Parameters}
To systematically investigate their impact on detection performance in rugby scenarios, we conducted experiments by varying key UAV operational parameters: fleet sizes (ranging from 4 to 20 UAVs), flight speeds (ranging from 0.1 to 11 m/s, incremented by 2 m/s), and detection radius (ranging from 3 to 8 m, incremented by 1 m). Based on preliminary experimental observations indicating significant variations in detection accuracy and identifiable stability thresholds at certain UAV fleet sizes, configurations were categorized into four distinct UAV-number groups: 4–7, 7–13, 13–16, and 16–20 UAVs. This grouping facilitates clear identification of performance improvements or stability issues associated with scaling the UAV fleet.

   \begin{figure}[htbp]
	\centering
	\begin{minipage}{0.49\linewidth}
		\centering
		\includegraphics[width=1.13 \linewidth]{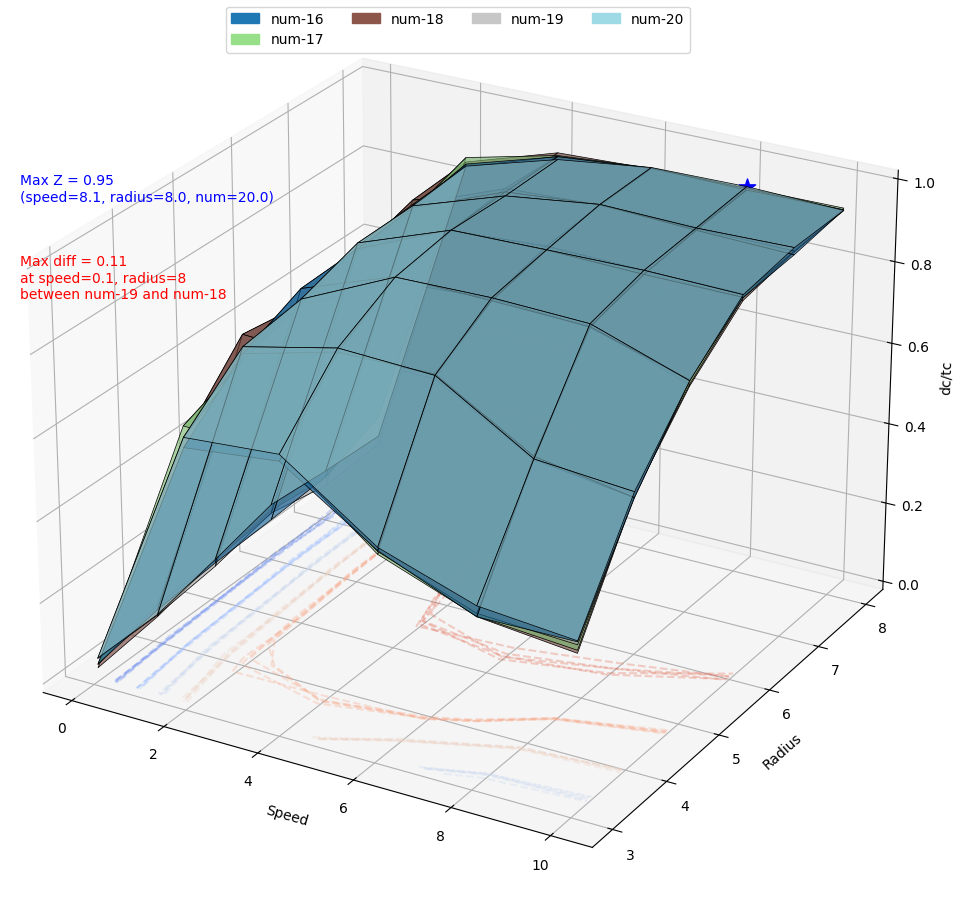}
		\caption{fp-N16-20}
		\label{fp-N16-20}
	\end{minipage}
	\begin{minipage}{0.49\linewidth}
		\centering
		\includegraphics[width=1.13 \linewidth]{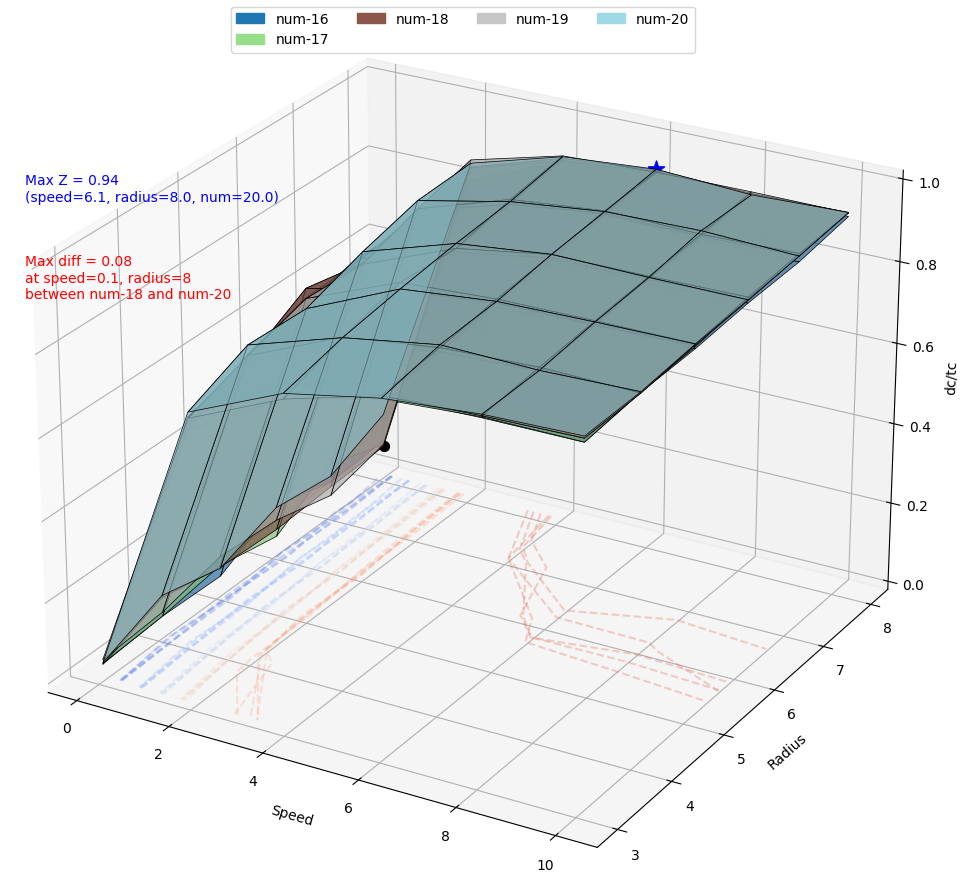}
		\caption{Density-N16-20}
		\label{Density-N16-20}
	\end{minipage}
  \end{figure}

   \begin{figure}[htbp]
	\centering
	\begin{minipage}{0.49\linewidth}
		\centering
		\includegraphics[width=1.13 \linewidth]{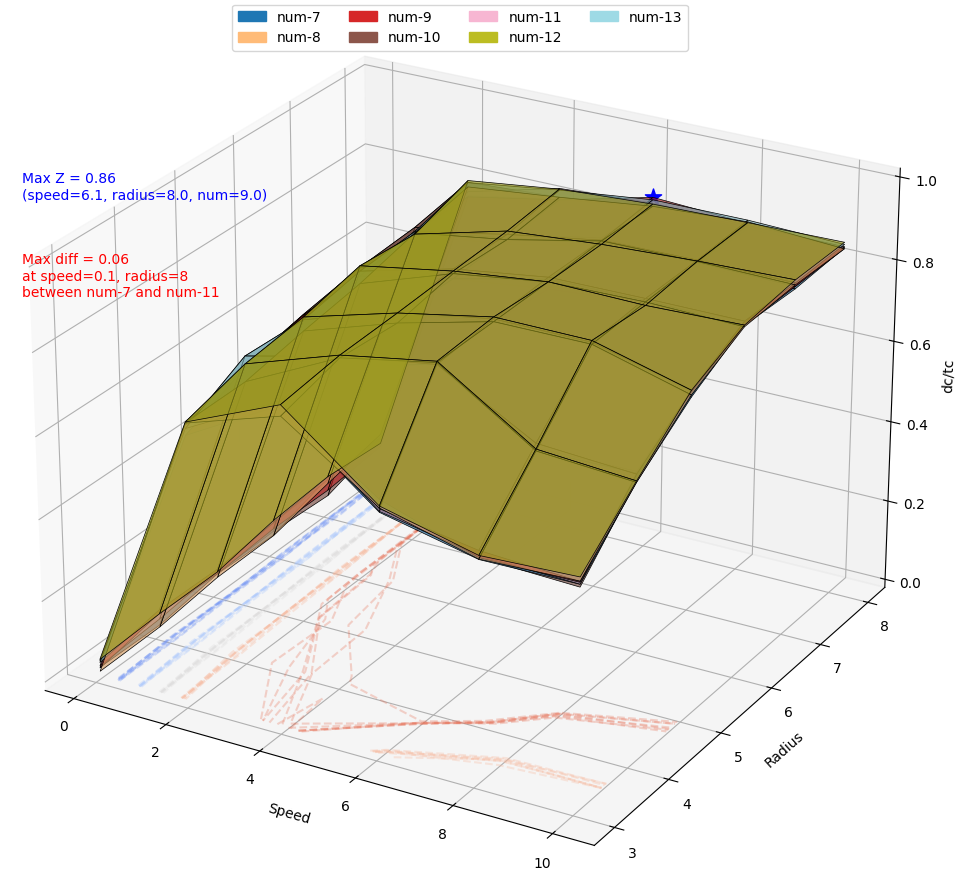}
		\caption{fb-N7-13}
		\label{fb-N7-13}
	\end{minipage}
	\begin{minipage}{0.49\linewidth}
		\centering
		\includegraphics[width=1.13 \linewidth]{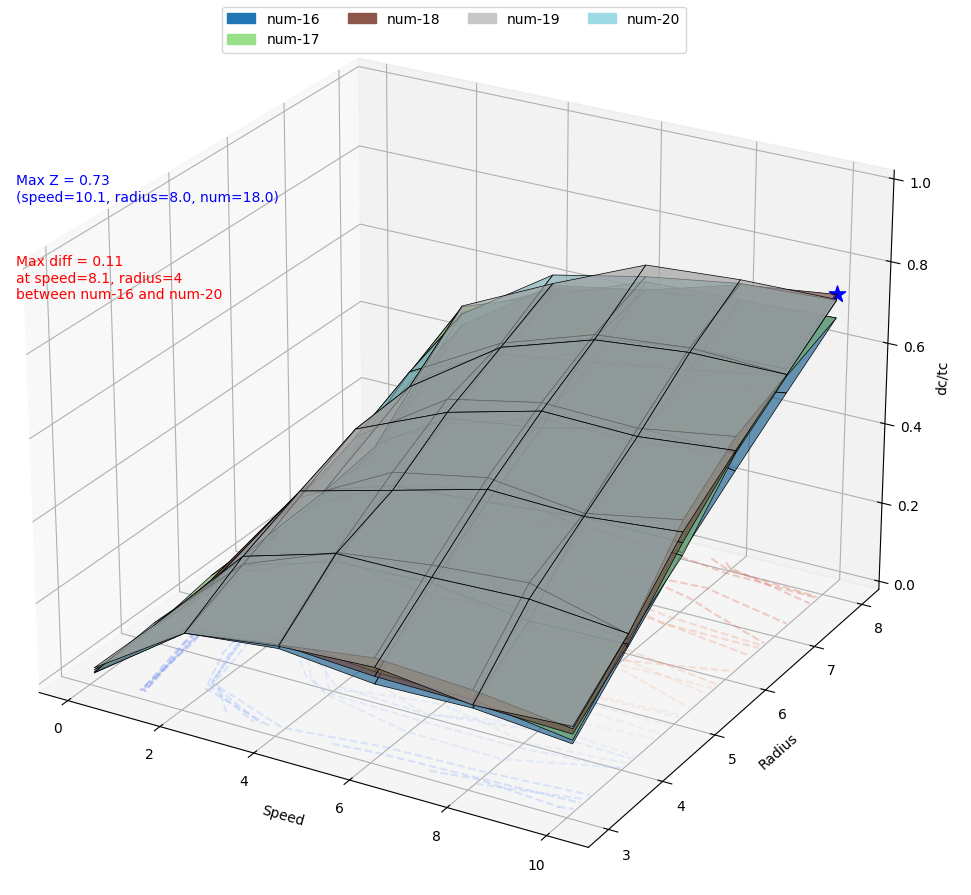}
		\caption{Random-N16-20}
		\label{Random-N16-20}
	\end{minipage}
        \caption*{Comparison of detection accuracy ($dc/rc$) among different UAV strategies. Although each subfigure shows results for different UAV counts, all accuracy measurements were conducted under consistent parameter variations, with detection radius ranging from 3 to 8 (step size = 1) and flight speed ranging from 0.1 to 11 (step size = 2).}
   \end{figure}

Fig.\ref{fp-N16-20} and Fig.\ref{Density-N16-20} clearly demonstrate that the dynamic strategies, specifically \textbf{\texttt{Follow-players}} and \textbf{\texttt{Density-based}}, achieved the highest detection accuracies across all tested configurations. The \textbf{\texttt{Follow-players}} strategy reached a peak accuracy of 0.95 at 20 UAVs (speed = 8.1, radius = 8), closely followed by the \textbf{\texttt{Density-based}} strategy, which peaked at 0.94 under similar conditions (20 UAVs, speed = 6.1, radius = 8). This indicates that dynamically tracking individual players or player-density zones significantly enhances collision detection capabilities.

Conversely, Fig.\ref{fb-N7-13} illustrates the moderate yet stable performance of the \textbf{\texttt{Follow-ball}} strategy, achieving an accuracy of approximately 0.86, independent of UAV count increments. However, Fig.\ref{Random-N16-20} demonstrates notably lower performance for the \textbf{\texttt{Random}} strategy, peaking only at 0.73. This underscores the performance limitations when UAV paths lack strategic targeting.

    \begin{table}[htbp]
    \centering
    \caption{Maximum Accuracy Achieved by Different Drone Strategies and Configurations}
    \begin{tabular}{cccccc}
        \hline
        Strategy & Drone Group & Drones & Speed & Radius & Accuracy \\ \hline
        Density & 4-7 & 7 & 10.1 & 8 & 0.88 \\ 
        Density & 7-13 & 13 & 10.1 & 8 & 0.92 \\ 
        Density & 13-16 & 14 & 6.1 & 8 & 0.93 \\ 
        Density & 16-20 & 20 & 6.1 & 8 & 0.94 \\ \hline
        Follow-ball & 4-7 & 6 & 10.1 & 8 & 0.86 \\ 
        Follow-ball & 7-13 & 9 & 6.1 & 8 & 0.86 \\ 
        Follow-ball & 13-16 & 16 & 10.1 & 8 & 0.86 \\ 
        Follow-ball & 16-20 & 19 & 8.1 & 8 & 0.87 \\ \hline
        Fixed & 4-7 & 7 & 6.1 & 8 & 0.51 \\ 
        Fixed & 7-13 & 10 & 10.1 & 8 & 0.54 \\ 
        Fixed & 13-16 & 15 & 6.1 & 8 & 0.53 \\ 
        Fixed & 16-20 & 19 & 10.1 & 8 & 0.55 \\ \hline
        Follow-players & 4-7 & 7 & 6.1 & 8 & 0.94 \\ 
        Follow-players & 7-13 & 11 & 6.1 & 8 & 0.94 \\ 
        Follow-players & 13-16 & 16 & 8.1 & 8 & 0.94 \\ 
        Follow-players & 16-20 & 20 & 8.1 & 8 & 0.95 \\ \hline
        Random & 4-7 & 7 & 8.1 & 8 & 0.46 \\ 
        Random & 7-13 & 13 & 10.1 & 8 & 0.62 \\ 
        Random & 13-16 & 16 & 10.1 & 8 & 0.67 \\ 
        Random & 16-20 & 18 & 10.1 & 8 & 0.73 \\ \hline
        Repulsive & 4-7 & 7 & 10.1 & 5 & 0.46 \\ 
        Repulsive & 7-13 & 13 & 2.1 & 8 & 0.60 \\ 
        Repulsive & 13-16 & 16 & 2.1 & 8 & 0.62 \\ 
        Repulsive & 16-20 & 19 & 2.1 & 8 & 0.69 \\ \hline
    \end{tabular}
    \label{tab:drone_strategy_results}
\end{table}

Detailed maximum accuracy values achieved under different UAV configurations for each strategy are summarized in Table~\ref{tab:drone_strategy_results}. The data clearly reinforce the advantage of \textbf{\texttt{Follow-players}} and \textbf{\texttt{Density-based}} approaches, consistently outperforming other strategies. Conversely, \textbf{\texttt{Fixed}} and \textbf{\texttt{Random}} strategies exhibit limited maximum accuracies of 0.55 and 0.73 respectively, highlighting their ineffectiveness in dynamic collision detection scenarios.

\begin{table}[htbp]
\centering
\caption{Maximum Accuracy Errors Across UAV Strategies and Configurations}
\begin{tabular}{|l|c|c|c|c|}
\hline
\textbf{Strategy} & \textbf{UAVs} & \textbf{Max Error} & \begin{tabular}[c]{@{}c@{}}Speed m/s\\ Radius m\end{tabular} & \begin{tabular}[c]{@{}c@{}}UAV\\ No. Compared\end{tabular} \\ \hline

\multirow{4}*{Density-based}
& 4–7    & 0.13 & 2.1, 3 & 5 vs. 7 \\
& 7–13   & 0.14 & 2.1, 3 & 7 vs. 13 \\
& 13–16  & 0.11 & 0.1, 8 & 14 vs. 15 \\
& 16–20  & 0.08 & 0.1, 8 & 18 vs. 20 \\ \hline

\multirow {4}*{Follow-ball}
& 4–7    & 0.06 & 2.1, 5 & 4 vs. 7 \\
& 7–13   & 0.06 & 0.1, 8 & 7 vs. 11 \\
& 13–16  & 0.07 & 0.1, 8 & 13 vs. 15 \\
& 16–20  & 0.06 & 0.1, 8 & 18 vs. 17 \\ \hline

\multirow {4}*{Fixed }
& 4–7    & 0.19 & 6.1, 8 & 4 vs. 7 \\
& 7–13   & 0.21 & 6.1, 7 & 11 vs. 9 \\
& 13–16  & 0.09 & 8.1, 8 & 13 vs. 16 \\
& 16–20  & 0.11 & 8.1, 8 & 18 vs. 17 \\ \hline

\multirow {4}*{Follow-players}
& 4–7    & 0.06 & 0.1, 8 & 4 vs. 7 \\
& 7–13   & 0.09 & 0.1, 6 & 7 vs. 13 \\
& 13–16  & 0.07 & 0.1, 8 & 13 vs. 15 \\
& 16–20  & 0.11 & 0.1, 8 & 19 vs. 18 \\ \hline

\multirow {4}*{Random}
& 4–7    & 0.21 & 10.1, 8 & 4 vs. 7 \\
& 7–13   & 0.21 & 10.1, 8 & 7 vs. 13 \\
& 13–16  & 0.11 & 10.1, 6 & 13 vs. 15 \\
& 16–20  & 0.11 & 8.1, 4  & 16 vs. 20 \\ \hline

\multirow {4}*{Repulsive}
& 4–7    & 0.18 & 2.1, 7 & 5 vs. 7 \\
& 7–13   & 0.28 & 2.1, 8 & 7 vs. 13 \\
& 13–16  & 0.12 & 0.1, 8 & 13 vs. 14 \\
& 16–20  & 0.12 & 0.1, 8 & 13 vs. 14 \\ \hline
\end{tabular}
\label{tab:max_error}
\end{table}

Table~\ref{tab:max_error} presents the maximum accuracy errors observed across UAV strategies. Notably, the \textbf{\texttt{Follow-ball}} and \textbf{\texttt{Follow-players}} strategies demonstrate minimal maximum errors (ranging from 0.06 to 0.11), suggesting robust and reliable performance even under suboptimal configurations. In contrast, the \textbf{\texttt{Fixed}} (up to 0.21) and \textbf{\texttt{Random}} (up to 0.21) strategies exhibit high variability, indicating unstable and unpredictable outcomes when UAV number or parameters vary.

In summary, these experimental results clearly demonstrate that UAV strategies incorporating dynamic target prioritization, particularly \textbf{\texttt{Follow-players}} and \textbf{\texttt{Density-based}} are superior both in achieving high detection accuracy and maintaining stability. These findings suggest that the targeted allocation of UAV resources toward key players or high-density player regions provides significant advantages over static or non-strategic deployment methods.

Based on the comprehensive analysis presented earlier, it was clear that UAV detection accuracy significantly varied with UAV strategies, speed, radius, and the number of UAVs. To further investigate how the accuracy is specifically affected by changes in the UAV fleet size under fixed operational conditions, several representative scenarios were selected to specifically examine the influence of UAV quantity under fixed radius and speed conditions.

\begin{itemize}
    \item Scenario 1 (Radius=8, Speed=8.1 m/s):
    As illustrated previously (Fig.~\ref{fp-N16-20}), the \textbf{\texttt{Follow-players}} strategy achieved maximum accuracy (0.95) at 20 UAVs. At this condition, varying UAV numbers clearly demonstrate the incremental advantage of deploying additional UAVs. For example, at 16 UAVs accuracy was 0.94, reflecting a marginal yet meaningful improvement when scaled to 20 UAVs, thus highlighting the potential benefit-to-cost trade-off.
    \item Scenario 2 (Radius=8, Speed=6.1 m/s):
    Under this scenario, the \textbf{\texttt{Density-based}} strategy yielded its best performance (0.94 at 20 UAVs). Comparative analysis of smaller UAV groups (e.g., accuracy = 0.93 at 14 UAVs) indicates diminishing returns after a certain UAV count threshold. This informs resource allocation decisions, suggesting that beyond approximately 14 UAVs, additional UAV deployment results in minimal performance gains.
    \item Scenario 3 (Radius=5, Speed=10.1 m/s):
    The \textbf{\texttt{Repulsive}} and \textbf{\texttt{Follow-ball}} strategies showed limited accuracies (0.46 and 0.86, respectively). This clearly demonstrates a challenging operating condition. However, comparing UAV counts within these conditions, even small increases in UAV number notably stabilize detection accuracy. For example, transitioning from 5 to 7 UAVs, accuracy for \textbf{\texttt{Repulsive}} strategies stabilizes significantly, reducing fluctuations and suggesting that at least 7 UAVs might be required to achieve reliable results under constrained radius conditions.
    \item Scenario 4 (Radius=3, Speed=2.1 m/s):
    This configuration exposed the largest accuracy errors in \textbf{\texttt{Density-based}} tracking (max error = 0.14), suggesting instability at low radius and speed conditions.
\end{itemize}

\subsection{Impact of UAV Fleet Size on Detection Accuracy with Fixed Flight Speed and Detection Radius}
Subsequently, we investigate the specific effect of varying UAV fleet size on detection accuracy for each strategy under fixed flight speed and detection radius conditions. In this part of the study, we evaluated detection performance for four representative scenarios:
 \begin{itemize}
     \item Scenario 1 (Radius=8, Speed=8.1 m/s)
     \item Scenario 2 (Radius=8, Speed=6.1 m/s)
     \item Scenario 3 (Radius=5, Speed=10.1 m/s)
     \item Scenario 4 (Radius=3, Speed=2.1 m/s)
\end{itemize}

For each scenario, we measured the detection accuracy of various strategies (\textbf{\texttt{Density-based}}, \textbf{\texttt{Follow-ball}}, \textbf{\texttt{Fixed (Heat Map)}}, \textbf{\texttt{Follow-players}}, \textbf{\texttt{Random}}, and \textbf{\texttt{Repulsive}}) across UAV swarm sizes ranging from 1 to 35 drones. 

\begin{figure}[htbp]
	\centering
	\begin{minipage}{0.49\linewidth}
		\centering
		\includegraphics[width=1.13\linewidth]{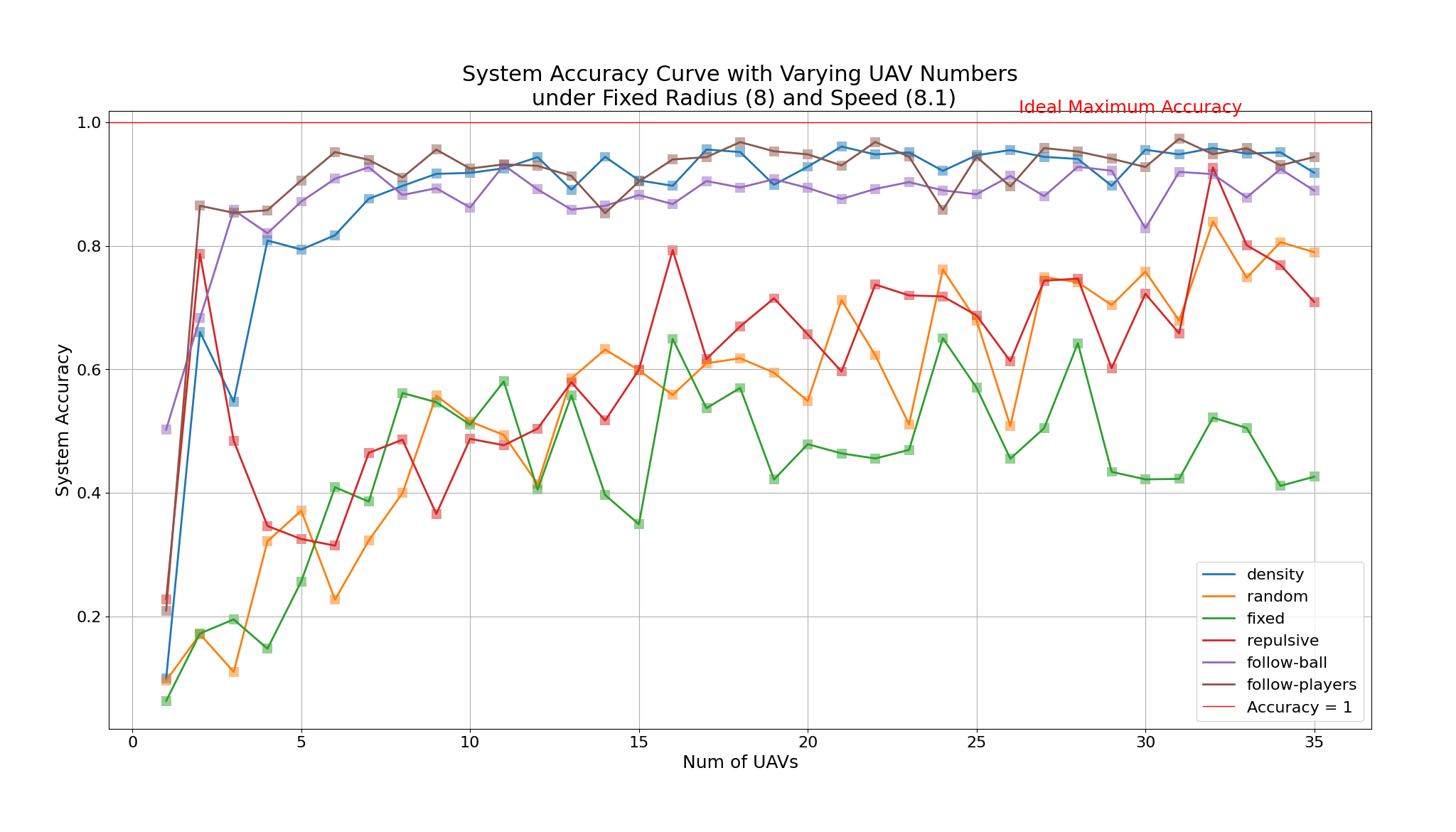}
		\caption{Scenario 1 (Radius=8, Speed=8.1 m/s)}
		\label{r8s8.1}
	\end{minipage}
	\hfill
	\begin{minipage}{0.49\linewidth}
		\centering
		\includegraphics[width=1.13\linewidth]{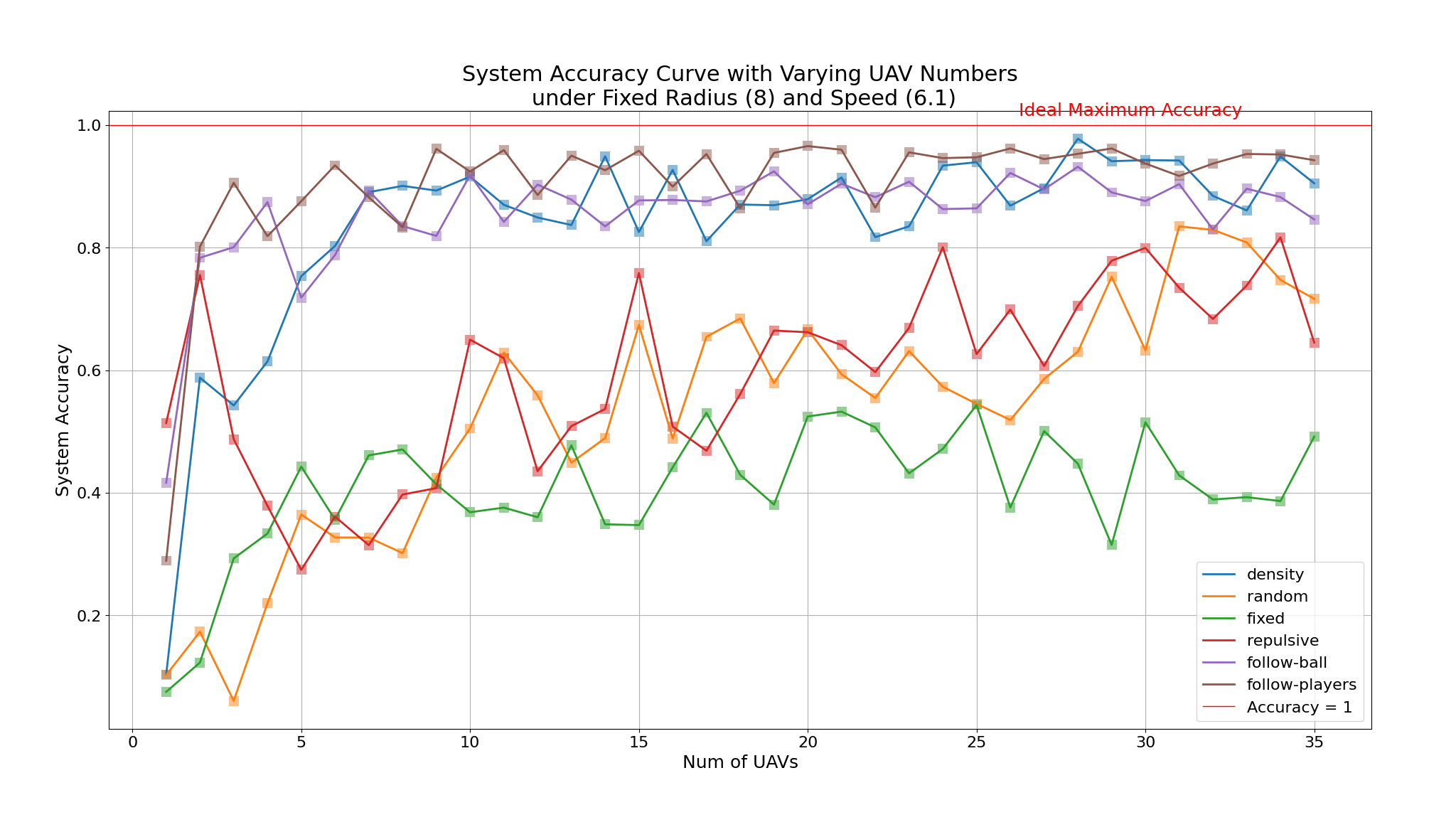}
		\caption{Scenario 2 (Radius=8, Speed=6.1 m/s)}
		\label{r8s6.1}
	\end{minipage}
	\vspace{1em} 
	\begin{minipage}{0.49\linewidth}
		\centering
		\includegraphics[width=1.13\linewidth]{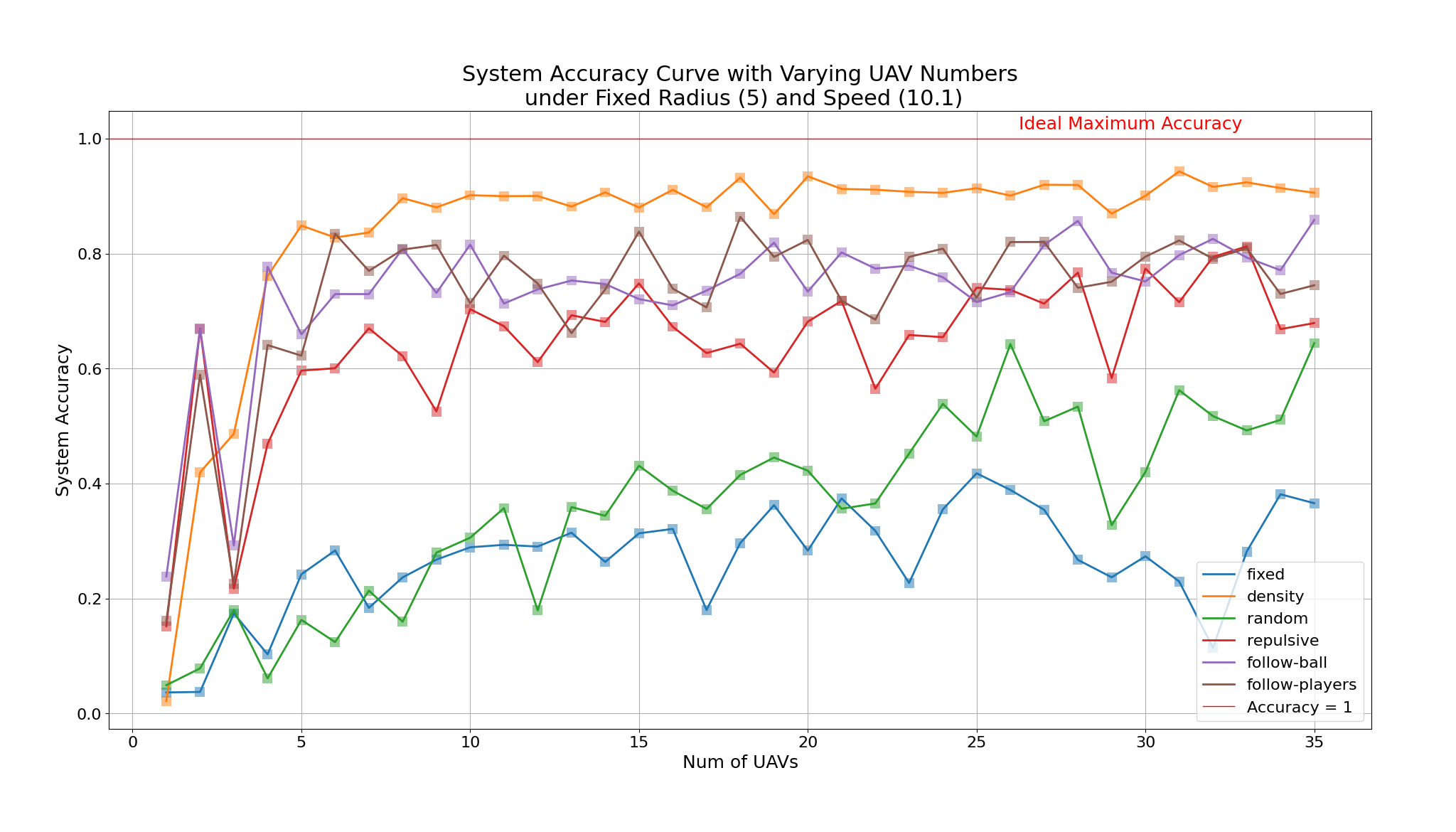}
		\caption{Scenario 3 (Radius=5, Speed=10.1 m/s)}
		\label{r5s10.1}
	\end{minipage}
	\hfill
	\begin{minipage}{0.49\linewidth}
		\centering
		\includegraphics[width=1.13\linewidth]{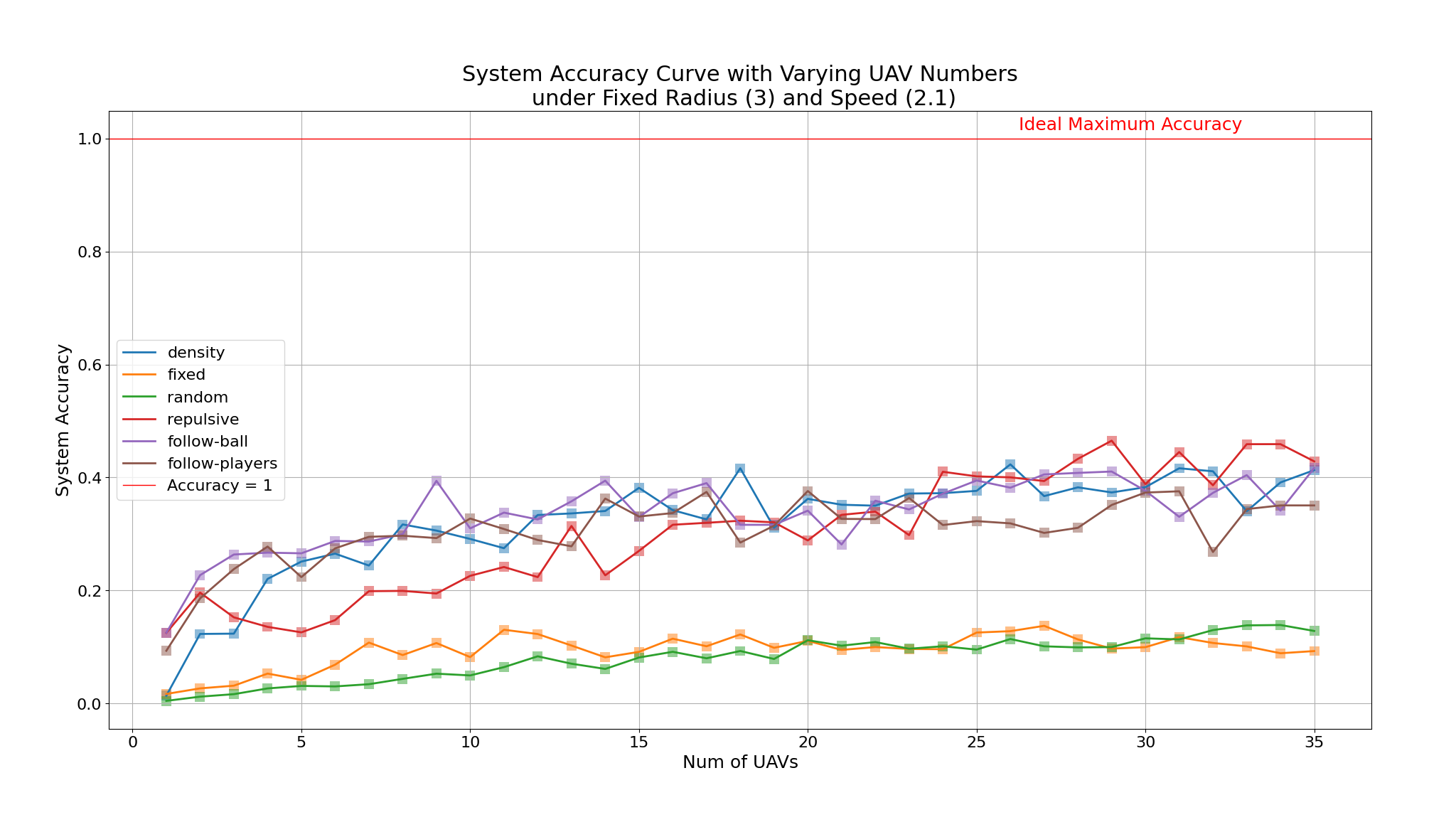}
		\caption{Scenario 4 (Radius=3, Speed=2.1 m/s)}
		\label{r3s2.1}
	\end{minipage}
\end{figure}

The Figure\ref{r8s8.1}, Figure\ref{r8s6.1}, Figure\ref{r5s10.1}, Figure\ref{r3s2.1} indicate that under fixed flight speed and detection radius, the detection accuracy improves as the UAV swarm size increases. In scenarios with larger radius and higher speeds (Scenarios 1 and 2), optimal performance is achieved with fewer UAVs, and the differences between strategies narrow as the swarm grows. Scenario 3, with a moderate radius and very high speed, demonstrates that while \textbf{\texttt{Follow-ball}} is best at very low UAV counts, the optimal strategy shifts to \textbf{\texttt{Follow-players}} for moderate numbers, with \textbf{\texttt{Density-based}} strategies ultimately outperforming as the swarm size increases. Conversely, in scenarios with small radius and slow speeds (Scenario 4), the benefit of adding additional UAVs is most pronounced, yet the optimal strategy transitions from \textbf{\texttt{Follow-ball}} to \textbf{\texttt{Follow-players}} and ultimately to \textbf{\texttt{Density-based}} or \textbf{\texttt{Repulsive}} modes at larger swarm sizes.

\begin{table}[htbp]
\centering
\caption{Selected Maximum Detection Accuracy for Best-Performing Strategies at Key UAV Counts}
\label{tab:scenario_accuracy}
\begin{tabular}{|c|c|c|c|}
\hline
\textbf{Scenario} & \textbf{UAV No.} & \begin{tabular}[c]{@{}c@{}}Optimal \\ Strategy \end{tabular} & \textbf{Accuracy} \\ \hline
1 (R=3, S=2.1) & 1--3 & Follow-ball & \begin{tabular}[c]{@{}c@{}}0.1251 (1 UAV)\\ 0.2271 (2 UAVs) \\ 0.2636 (3 UAVs)\end{tabular}\\ \hline
1 (R=8.1, S=8.1) & 1 & Follow-ball & 0.5026 \\ \hline
1 (R=8.1, S=8.1) & 3 & \begin{tabular}[c]{@{}c@{}}Follow-ball/ \\ Follow-players \end{tabular} & \begin{tabular}[c]{@{}c@{}}0.8579 (Follow-ball)\\ within 0.0044 of \\Follow-players \end{tabular}\\ \hline
1 (R=8.1, S=8.1) & 12--35 & \begin{tabular}[c]{@{}c@{}}Follow-players/ \\ Density-based \end{tabular} & 0.91--0.97 \\ \hline
2 (R=8, S=6.1) & 1 & Repulsive & 0.5138 \\ \hline
2 (R=8, S=6.1) & 3 & Follow-players & 0.9060 \\ \hline
2 (R=8, S=6.1) & 28 & Density-based & 0.9780 \\ \hline
3 (R=5, S=10.1) & 1 & Follow-ball & 0.2384 \\ \hline
3 (R=5, S=10.1) & 2 & Repulsive & 0.6694 \\ \hline
3 (R=5, S=10.1) & 3 & Density & 0.4868 \\ \hline
3 (R=5, S=10.1) & 4 & Follow-ball & 0.7771 \\ \hline
3 (R=5, S=10.1) & 5+ & Density & 0.8488--0.9432 \\ \hline
4 (R=3, S=2.1) & 4--10 & Follow-players & \begin{tabular}[c]{@{}c@{}}0.2777 (4 UAVs)\\ 0.3273 (10 UAVs)\end{tabular} \\ \hline
4 (R=3, S=2.1) & 29 & Repulsive & 0.4653 \\ \hline

\end{tabular}
\end{table}

 \begin{itemize}
     \item Scenario 1 (Radius=8, Speed=8.1 m/s)
In Scenario 1 (R=8.1 m, S=8.1 m/s), a single UAV performed best in \textbf{\texttt{Follow-ball}} mode (accuracy = 0.5026). As the number of UAVs increased, \textbf{\texttt{Follow-players}} and subsequently \textbf{\texttt{Density-based}} strategies dominated, with overall high detection rates (up to 0.97) for larger swarms.
     \item Scenario 2 (Radius=8, Speed=6.1 m/s)
In Scenario 2 (R=8 m, S=6.1 m/s), the best performance shifted from \textbf{\texttt{Repulsive}} for a single drone (0.5138) to \textbf{\texttt{Follow-players}} and \textbf{\texttt{Follow-ball}} for small to medium swarms, with \textbf{\texttt{Density-based}} becoming optimal for certain swarm sizes (e.g., peaking at 0.9780 for 28 drones).
     \item Scenario 3 (Radius=5, Speed=10.1 m/s)
In Scenario 3 (R=5 m, S=10.1 m/s), a single UAV again favored \textbf{\texttt{Follow-ball}} (accuracy = 0.2384), while for 2–3 UAVs the optimal mode shifted (e.g., \texttt{Repulsive} or \texttt{Density} modes), and beyond 4 drones, \textbf{\texttt{Follow-ball}} initially provided the highest performance before \textbf{\texttt{Density-based}} strategies became dominant in larger swarms.
     \item Scenario 4 (Radius=3, Speed=2.1 m/s)
In Scenario 4 (R=3 m, S=2.1 m/s), at small swarm sizes (1–3 drones), the \textbf{\texttt{Follow-ball}} mode consistently achieved the highest accuracy (e.g., 0.1251 for 1 drone, 0.2271 for 2 drones, and 0.2636 for 3 drones). As the swarm size increased, the optimal control mode transitioned: \textbf{\texttt{Follow-players}} became competitive for intermediate sizes (e.g., 0.2777 for 4 drones and 0.3273 for 10 drones), and for large swarms (around 29 drones), the \textbf{\texttt{Repulsive mode}} reached the highest accuracy (0.4653).
\end{itemize}

To further clarify the performance differences, Table~\ref{tab:scenario_accuracy} summarizes the best-performing strategy and its corresponding detection accuracy at selected UAV swarm sizes for each scenario. These data illustrate the transition points (e.g., from \texttt{Follow-ball} to \texttt{Follow-players}, and then to \texttt{Density-based} or \texttt{Repulsive} modes) and the magnitude of performance differences between the optimal and suboptimal strategies.

\begin{table}[htbp]
\centering
\caption{Transition Points of Optimal Strategy Across Scenarios}
\label{tab:strategy_transitions}
\begin{tabular}{|c|c|c|}
\hline
\textbf{Scenario} & \begin{tabular}[c]{@{}c@{}}Transition from\\ Follow-ball to \\ Follow-players\end{tabular} & \begin{tabular}[c]{@{}c@{}}Transition to\\ Coverage-Based \\ (Density/Repulsive)\end{tabular} \\ \hline
1 (R=8.1, S=8.1) & $\approx$ 2--3 UAVs & $\approx$ 6--8 UAVs \\ \hline
2 (R=8, S=6.1) & $\approx$ 2--3 UAVs & $\approx$ 8--10 UAVs \\ \hline
3 (R=5, S=10.1) & $\approx$ 3--5 UAVs & $\approx$ 15--20 UAVs \\ \hline
4 (R=3, S=2.1) & $\approx$ 3--4 UAVs & $\geq$ 20 UAVs \\ \hline
\end{tabular}
\end{table}

Comparing across scenarios reveals that larger detection radius and higher UAV speeds generally yield higher overall accuracies and can shift the optimal strategy thresholds to lower UAV counts. For instance, while Scenario 4 (with a small R) required a much larger swarm to achieve high accuracy, Scenarios 1 and 2 reached near-saturation performance at moderate swarm sizes (10–15 UAVs). In Scenario 3, despite the very high speed, the moderate sensor range necessitated more drones to achieve complete coverage. Table~\ref{tab:strategy_transitions} provides an overview of the transition points for the optimal strategy as a function of swarm size in each scenario, while Table~\ref{tab:performance_gap} details the performance gaps (difference between best and worst strategies) at representative UAV counts.

\begin{table}[htbp]
\centering
\caption{Representative Performance Gaps (Accuracy Difference Between Best and Worst Strategies)}
\label{tab:performance_gap}
\begin{tabular}{|c|c|c|}
\hline
\textbf{Scenario} & \textbf{UAV Count} & \textbf{Accuracy Gap (Best - Worst)} \\ \hline
1 (R=8.1, S=8.1) & 3 UAVs & $\sim$ 0.0044 (between top strategies) \\ \hline
2 (R=8, S=6.1) & 1 UAV & $\sim$ 0.4388 \\ \hline
2 (R=8, S=6.1) & 29 UAVs & $\sim$ 0.6471 \\ \hline
3 (R=5, S=10.1) & 1 UAV & $\sim$ 0.2163 \\ \hline
3 (R=5, S=10.1) & 4 UAVs & $\sim$ 0.7159 \\ \hline
4 (R=3, S=2.1) & 1 UAV & $\sim$ 0.1206 \\ \hline
4 (R=3, S=2.1) & 10 UAVs & $\sim$ 0.15--0.20 \\ \hline
\end{tabular}
\end{table}

Based on the results presented in Tables~\ref{tab:scenario_accuracy}, \ref{tab:strategy_transitions}, and \ref{tab:performance_gap}, several key observations can guide the selection of fixed UAV numbers for subsequent experiments. This selection aims to facilitate a detailed evaluation of how variations in detection radius independently affect collision detection accuracy, keeping UAV count and flight speed constant.

Analysis of Table~\ref{tab:scenario_accuracy} indicates that detection accuracy generally stabilizes or reaches saturation within medium-to-large UAV swarms (approximately 10–20 UAVs) for scenarios involving higher detection radius (e.g., \(R \geq 8\) m). For instance, in Scenario 2 (\(R=8\) m, \(S=6.1\) m/s), peak accuracy (0.9780) occurs with 28 UAVs, but similarly high accuracy (\(\geq 0.94\)) is consistently achieved at approximately 10–12 UAVs in Scenario 1 (\(R=8.1\) m, \(S=8.1\) m/s). In contrast, smaller swarm sizes (1–5 UAVs) exhibited greater variability in optimal strategy selection and pronounced performance gaps, complicating clear isolation of radius effects. Table~\ref{tab:strategy_transitions} further emphasizes that critical strategy transitions, from \textbf{\texttt{Follow-ball}} to \textbf{\texttt{Follow-players}} and subsequently to \textbf{\texttt{Density-based}} or \textbf{\texttt{Repulsive modes}}, predominantly occur between 8 and 20 UAVs. This range provides a balanced context in which shifts in strategy dominance due to variations in detection radius can be clearly and meaningfully analyzed. Moreover, Table~\ref{tab:performance_gap} highlights that substantial performance differences (up to 0.7159) emerge at lower UAV counts, indicative of unstable conditions where accuracy is overly sensitive to strategic choice. Conversely, at intermediate swarm sizes (approximately 10–15 UAVs), the performance gaps between optimal and suboptimal strategies become moderate and relatively stable, making it easier to detect nuanced changes resulting from adjustments in detection radius alone.

Based on the aforementioned results, we adopt a fixed UAV fleet size of 12 drones for subsequent experimentation. This fleet size was selected due to its ability to consistently achieve high detection accuracy (\(\geq 0.9\)) using \textbf{\texttt{Follow-players}} or \textbf{\texttt{Density-based}} strategies. Additionally, a fleet of 12 drones lies within the critical transitional range (8–20 UAVs), enabling clear observation of radius-dependent shifts in strategy effectiveness. Moderate and meaningful performance gaps observed at this UAV number further justify its selection, facilitating precise analysis of how sensor range variations independently influence strategic performance.

\subsection{Impact of Detection Radius on Accuracy with Constant UAV Fleet Size and Flight Speed}

In this subsection, we specifically investigate how varying the detection radius affects collision detection accuracy under controlled conditions—maintaining a constant UAV fleet size and flight speed. This targeted analysis isolates the role of sensor range in determining strategic effectiveness and collision detection capability.

The experiments conducted herein adhere strictly to the following parameters:

\begin{itemize}
    \item \textbf{Fixed UAV fleet size:} 12 drones
    \item \textbf{Constant flight speed:}  8 m/s, selected based on stable, high-accuracy conditions identified in previous scenarios
    \item \textbf{Detection radius variation:} incrementally adjusted from 2 m to 15 m, at intervals of 0.5 m
\end{itemize}

  \begin{figure}[h]
   \centering
   \includegraphics[width=3.5in]{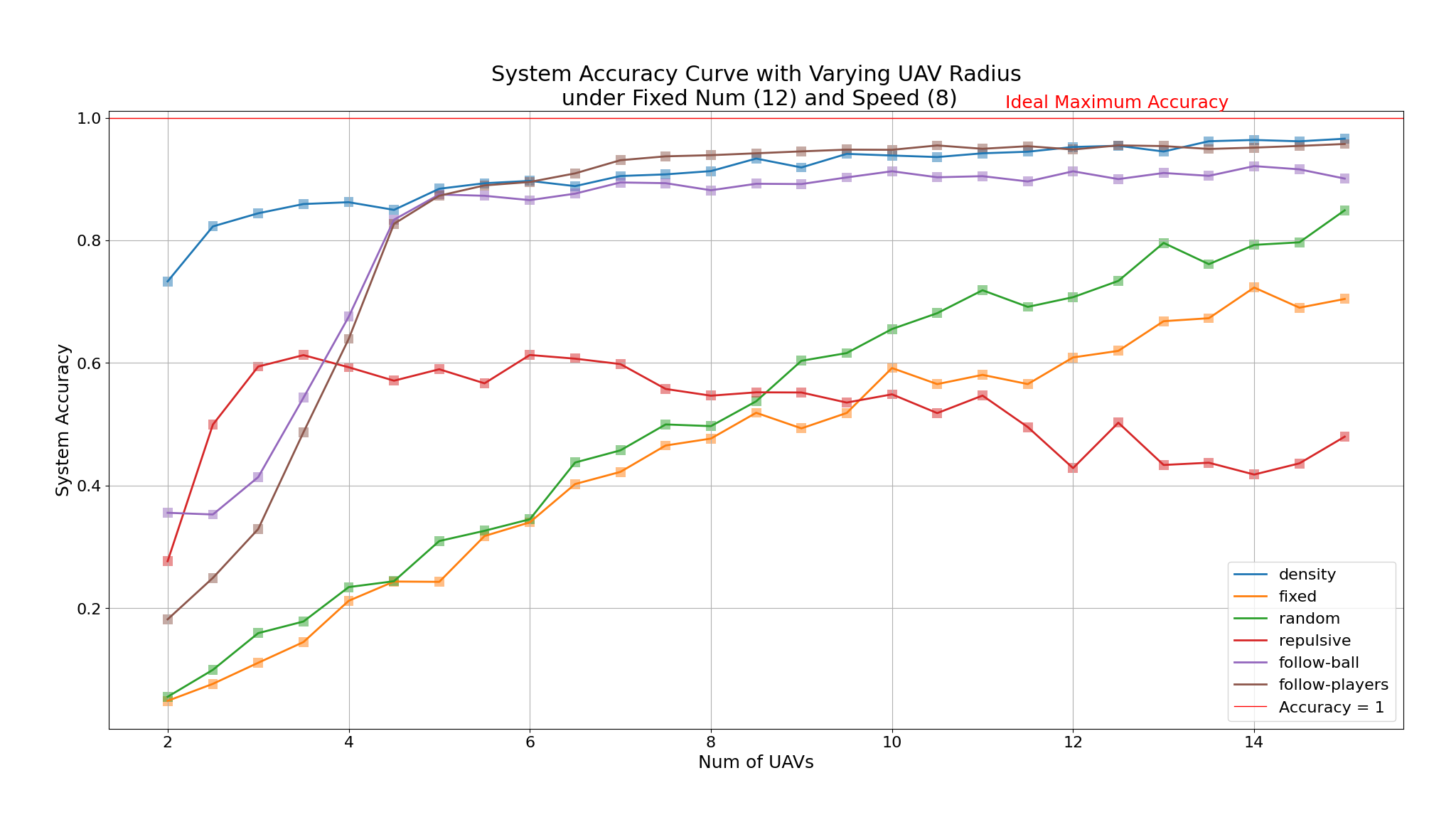}
   \caption{system performance of fixed radius and speed varying num}
   \label{fixed-num-v-radius}
   \end{figure}

The Figure\ref{fixed-num-v-radius} comprehensively demonstrate clear trends and key transition points for optimal strategy selection. At lower radius (2–4 m), the \textbf{\texttt{Density-based}} strategy consistently provided the highest accuracy, increasing sharply from 0.8228 at 2.5 m to 0.8623 at 4 m. The greatest performance gaps in this range consistently occurred between the \textbf{\texttt{Density-based}} and \textbf{\texttt{Fixed}} strategies, highlighting the significant advantage of dynamic targeting over static or random methods at constrained detection ranges. The minimal performance differences observed at small radius were typically between \textbf{\texttt{Random}} and \textbf{\texttt{Fixed}} strategies, indicating their similarly limited effectiveness.

Between moderate radius (4–6.5 m), accuracy continued to rise notably, reaching 0.8971 at 6.5 m. The \textbf{\texttt{Follow-players}} mode started to show competitive performance from approximately 5 m, eventually surpassing the \textbf{\texttt{Density-based}} strategy at 6.5 m (accuracy 0.9093). Performance gaps between optimal and lower-performing strategies (particularly \textbf{\texttt{Fixed}} and \textbf{\texttt{Random}}) increased substantially, highlighting the critical advantage of dynamic strategies as detection radius expands.

In the higher radius range (7–15 m), \textbf{\texttt{Follow-players}} became predominantly optimal, achieving peak accuracies between 0.9308 (at 7 m) and 0.9550 (at 12.5 m). \textbf{\texttt{Density-based}} strategy closely followed, maintaining marginally lower yet consistently competitive performance (around 0.9350 to 0.9638 at larger radius such as 14 m). The largest Performance gaps (up to approximately 0.5) consistently occurred between the optimal strategies (\textbf{\texttt{Follow-players}} or \textbf{\texttt{Density-based}}) and the \textbf{\texttt{Repulsive}} or \textbf{\texttt{Fixed}} modes, indicating significant disadvantages in non-adaptive strategies at these sensor ranges.

\begin{table}[ht]
\centering
\caption{Comparison of Strategies at Different Detection Radius}
\label{tab:strategy_comparison}
\begin{tabular}{|c|c|c|c|}
\hline
\textbf{Radius (m)} & \begin{tabular}[c]{@{}c@{}}Optimal \\ Strategy \end{tabular} & \textbf{Accuracy} & \begin{tabular}[c]{@{}c@{}}Performance \\ Gap\end{tabular} \\ \hline
2.0   & Density-based       & 0.7331   & \begin{tabular}[c]{@{}c@{}} $\approx$0.6842\\ (vs. Fixed/Random)\end{tabular} \\ \hline
6.5   & Follow-players      & 0.9093   & 0.5771 (vs. Fixed) \\ \hline
$>$10 & \begin{tabular}[c]{@{}c@{}}Follow-players/\\ Density-based\end{tabular}& -- & 0.001--0.009 \\ \hline
\end{tabular}
\end{table}

The Table\ref{tab:strategy_comparison} indicates that at small detection radius (e.g., 2 m), the \textbf{\texttt{density-based}} strategy significantly outperforms both the \textbf{\texttt{fixed}} and \textbf{\texttt{random}} strategies. As the sensor range increases to 6.5 m, the \textbf{\texttt{follow-players}} strategy becomes optimal, and beyond 10 m, the performance differences among the strategies become minimal, suggesting a degree of interchangeability under such conditions.

\section{Discussion}

The results demonstrate that decentralized UAV-based strategies can effectively detect collisions in rugby scenarios. Our simulation showed that a fleet of autonomous drones, each making local decisions and sharing data, achieved higher detection accuracy and responsiveness compared to a single drone or traditional fixed cameras. By coordinating their coverage, the drones reduced blind spots and occlusion issues, capturing collision events that a stationary viewpoint might miss. This led to notable accuracy improvements: as the number of UAVs increased or as their deployment became more dynamic, more collisions were detected in real time with fewer false negatives. These findings align with prior evidence that automated collision monitoring in rugby is feasible and can closely match expert video analysis.

Deploying a large number of drones in a real match, however, raises practical feasibility questions. While more drones can widen coverage, there are diminishing returns and added complexities when scaling up the fleet. Each additional UAV introduces coordination overhead and potential airspace conflicts, and there are limits to how many can be safely and legally deployed over a crowded venue. Regulatory frameworks impose strict rules on drone operations in public spaces; organizers must obtain flight permissions and ensure compliance with aviation laws. In practice, this means only a limited fleet (perhaps a handful of drones) could be realistically used during a live rugby match before the logistical, regulatory, and safety challenges outweigh the benefits.

Operation range poses another crucial consideration. Our experimental results show that UAV strategies exhibit varying effectiveness depending on operational parameters such as detection radius. At lower detection radius, more UAVs are necessary to achieve stable and accurate collision detection, making deployment less efficient in practical applications. Conversely, larger detection radius enhance performance considerably but may introduce challenges such as increased interference between drones, higher power consumption, and potential regulatory concerns due to broader surveillance coverage.

Another consideration is the quality of the camera sensors and their ability to capture high-speed collisions. Current UAV-mounted cameras are increasingly sophisticated, often supporting high-resolution and high-frame-rate video (e.g., 4K at 60 fps). In our experiments, we assumed these capabilities are sufficient to discern collision events and potential head impacts. For many scenarios, standard drone cameras do provide clear footage of tackles and impacts, especially with features like gimbal stabilization and high shutter speeds for daylight conditions. However, extremely fast impacts or subtle injury signs (like transient loss of consciousness) might still be missed if the frame rate or resolution isn’t high enough. In low-light or bad weather conditions, image quality could degrade, so ensuring cameras have good low-light performance or using thermal/IR sensors might be necessary in future iterations\cite{thermal01, thermal02}.

Operational challenges must also be addressed before UAV monitoring can be used in real games. Rugby is fast-paced, and drones need to adjust speed and position rapidly to keep players in frame. High-end drones can reach top speeds around 20–21 $m/s$ , which is on par with or faster than the sprinting players, so in theory they can keep up with play. Advanced flight modes and obstacle-sensing technology (e.g., vision-based tracking and collision avoidance) are already available, enabling drones to navigate complex, dynamic environments. Nonetheless, sudden direction changes, scrums, and mid-air contests pose a challenge for maintaining a stable view; a drone might need to predict player movements or smoothly circle around a maul to avoid losing line-of-sight. Limited flight time is another significant constraint: many UAVs can only fly about 15–30 minutes on a single battery charge\cite{jacapture}. Covering an entire 80-minute rugby match would require multiple drones taking turns or quick battery swaps at stoppages. While the radio range of modern drones (often several kilometers) is more than sufficient for a single field, coordinating multiple UAVs in the same airspace is non-trivial. Robust inter-drone communication and collision-avoidance protocols are needed to prevent drones from interfering with each other or the players. In summary, the decentralized UAV approach shows promise in accuracy and coverage, but real-world deployment will require careful consideration of hardware limits, safety protocols, and regulatory compliance.

\section{Conclusions and Future Work}
This paper introduced a decentralized UAV-based collision monitoring framework tailored for rugby scenarios, aiming to enhance the detection accuracy of high-impact events and mitigate risks associated with traumatic brain injuries (TBIs). Our decentralized UAV system demonstrated superior performance through various innovative strategies, particularly the Follow-players and Density-based modes, outperforming traditional static approaches. Through extensive simulations using the NetLogo platform, we systematically analyzed the effects of UAV fleet size, flight speed, and detection radius on collision detection accuracy. The findings provide critical insights into the optimal configurations and strategic deployment of UAV fleets for effective and timely monitoring of collision events.

Future research directions include extending the current two-dimensional simulation framework to UAV angles in three-dimensional environments, allowing for a more realistic representation of rugby matches and capturing collision events from multiple perspectives. Investigating multi-UAV collaborative strategies to simultaneously capture and analyze the same collision event from various angles will further enhance detection accuracy and robustness. 
\subsection{UAV Angles in 3D Environments and Real-world Considerations}

The experiments conducted in this study were based on a simplified two-dimensional (2D) simulation environment using NetLogo. This simplification restricts UAV and player movement to a two-dimensional plane, thereby abstracting away critical three-dimensional (3D) operational factors. In realistic environments or actual rugby matches, UAVs operate within a 3D space, significantly impacting their ability to detect head collisions due to angular positioning and potential visual occlusions.

Future research must therefore address the limitations posed by 2D simulation environments by examining UAV positioning and camera angles in 3D contexts. This involves exploring various UAV orientations, such as drones hovering directly above play or positioned at tilted angles relative to the field, and determining their respective impacts on the visibility of player collisions. A systematic assessment, either by extending current simulations to incorporate a 3D model or through controlled field experiments, will be essential to ascertain optimal UAV angles and altitudes. Such studies will help identify strategies that minimize visual occlusions and maximize the effectiveness of head impact detection using airborne camera systems.

Moreover, our current findings highlight the advantages of adaptive UAV strategies within a simplified 2D simulation. However, practical deployment in 3D real-world scenarios demands a more thorough analysis of UAV spatial arrangements. Thus, future work should specifically explore how UAV altitude adjustments, angular positioning (e.g., vertical overhead versus angled views), and multi-UAV coordination in capturing single collision events from multiple perspectives may further improve detection accuracy and practical reliability.

\subsection{Multi-UAV Perspective on the Same Collision}
In our current decentralized system, each UAV is assigned a distinct coverage area to maximize overall surveillance efficiency. An intriguing enhancement involves deploying multiple UAVs to observe the same collision event from various angles and altitudes, thereby enriching the analytical depth of the captured data. Implementing a multi-perspective UAV framework offers several potential advantages. For instance, if one UAV's line of sight is obstructed during a collision, another UAV positioned differently may maintain an unobstructed view, thereby mitigating occlusion issues. Integrating feeds from multiple UAVs can enhance the accuracy and reliability of collision detection systems. 

Future research should focus on developing methodologies to effectively fuse video data from multiple UAVs into a cohesive analytical framework. Moreover, multi-UAV monitoring enables dynamic adjustment of UAV positions based on real-time data analysis. For example, if a UAV detects a potential collision, it could signal other UAVs to adjust their positions for optimal coverage, thereby improving data capture quality. Incorporating reinforcement learning algorithms could further enhance this adaptive positioning, allowing UAVs to learn and predict optimal vantage points over time.

\bibliographystyle{IEEEtran}
\bibliography{ref}

\end{document}